\documentclass{article}

\usepackage[final]{arxiv}
\usepackage{graphicx}
\usepackage{epstopdf}
\usepackage{caption}

\usepackage[utf8]{inputenc} 
\usepackage[T1]{fontenc}    
\usepackage{hyperref}       
\usepackage{url}            
\usepackage{booktabs}       
\usepackage{amsfonts}       
\usepackage{nicefrac}       
\usepackage{microtype}      
\usepackage{setspace}       
\usepackage{amsmath}
\usepackage{amssymb}
\usepackage{adjustbox}
\usepackage{tablefootnote}
\usepackage[bottom]{footmisc}

\usepackage{algorithmic}
\usepackage{subcaption}
\usepackage[ruled]{algorithm2e}

\usepackage{natbib}
\title{A Contrast Based Feature Selection Algorithm for High-dimensional Data set in Machine Learning}

\author{
	Chunxu CAO \And Qiang ZHANG
}

\begin{document}
	
	
	\maketitle
	
	\begin{abstract}
		
		Feature selection is an important process in machine learning and knowledge discovery. By selecting the most informative features and eliminating irrelevant ones, the performance of learning algorithms can be improved and the extraction of meaningful patterns and insights from data can be facilitated.
		
		However, most existing feature selection methods, when applied to large datasets, encountered the bottleneck of high computation costs. 
		
		To address this problem, we propose a novel filter feature selection method, ContrastFS, which selects discriminative features based on the discrepancies features shown between different classes. We introduce a dimensionless quantity as a surrogate representation to summarize the distributional individuality of certain classes, based on this quantity we evaluate features and study the correlation among them. We validate effectiveness and efficiency of our approach on several widely studied benchmark datasets, results show that the new method performs favorably with negligible computation in comparison with other state-of-the-art feature selection methods.
		
	\end{abstract}	
	
	\section{Introduction}
	
	
	High-dimensional data generated from advanced data collection and exploratory data analysis techniques are prevalent in domains such as text analysis\cite{yang1997comparative}, computer vision \cite{bolon-canedo2020feature}, bioinformatics  \cite{saeys2007review} and business analytics \cite{konda2013feature}, it provides rich information for machine learning and deep learning models to achieve good performance \cite{goodfellow2016deep}. 
	
	Meanwhile, they also bring many challenges \cite{johnstone2009statistical,fan2006statistical}, one of them is the curse of dimensionality \cite{bellman1957dynamic}, in this famous phenomenon the number of samples required to estimate an arbitrary function with a given level of accuracy grows exponentially with respect to the dimensionality of input of the function. It also implies that the distance between any two points in a high-dimensional space becomes almost the same, making similarity search challenging. 
	Additionally, high-dimensional datasets result in high storage and computational cost for learning algorithms, and reduce the interpretability of the learned models. 
	
	
	In contrast to the rapidly increasing number of features in modern datasets, only a small subset of features in datasets is relevant to the output in many problems of interest \cite{lemhadri2021lassonet}. 
	Feature selection is a feasible approach to find this subset \cite{ruppert2004elements}, it selects features from the original feature set, thereby reducing the dimensionality, decreasing the time cost, enhancing the generalization and improving performance of machine learning models \cite{guyon2003introduction, cai2018feature, li2018feature}.
	In many cases, it's not the accuracy of a specific learning algorithm we are most interested in, but rather finding a meaningful feature subset, e.g., knowledge discovery. 
	Feature selection is also a valuable tool in this setting \cite{saeys2008robust, dessi2015similarity, li2017recent}, as it can uncover the most relevant and meaningful features \cite{hanchuanpeng2005feature}, and facilitate data visualization \cite{yang1999data}.
	
	
	Due to these advantages and the massive growth of data across many scientific disciplines, a large number of feature selection methods have emerged in the past decades. 
	These methods have found numerous applications in various domains where high-dimensional data are common \cite{guyon2003introduction, chandrashekar2014survey, li2018feature, cai2018feature, bolon-canedo2020feature, hancer2020survey, solorio-fernandez2022survey}. Feature selection has therefore attracted a lot of attention from researchers and practitioners in the field of data science, with many workshops focusing on the topic \cite{nie2010efficient}.
		\begin{figure}[!t]
		\centering
		\begin{subfigure}[b]{0.242\textwidth}
			\includegraphics[width=\textwidth]{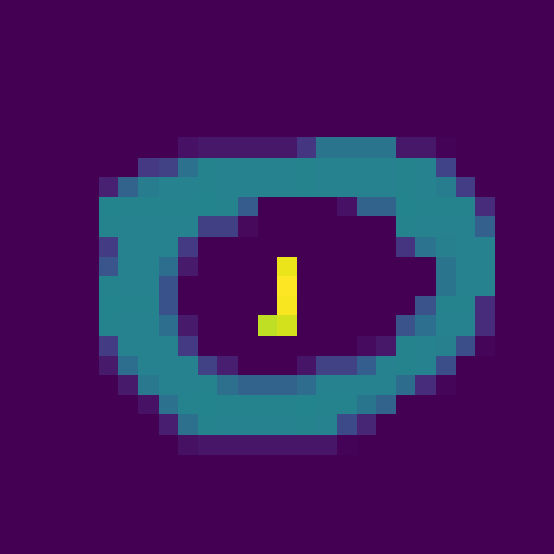}
			\caption{}
		\end{subfigure}
		\begin{subfigure}[b]{0.242\textwidth}
			\includegraphics[width=\textwidth]{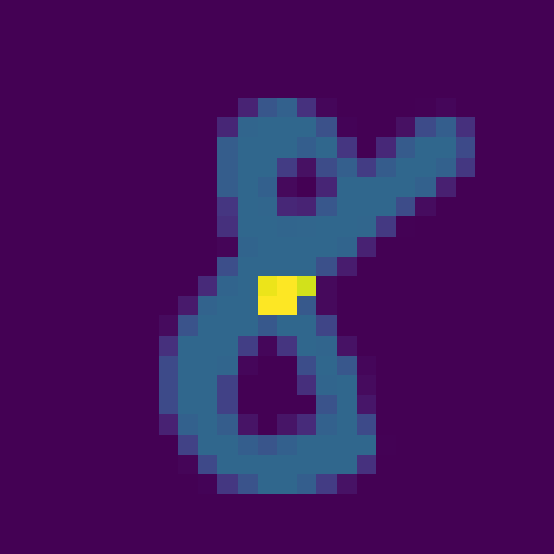}
			\caption{}
		\end{subfigure}
		~ 
		\begin{subfigure}[b]{0.242\textwidth}
			\includegraphics[width=\textwidth]{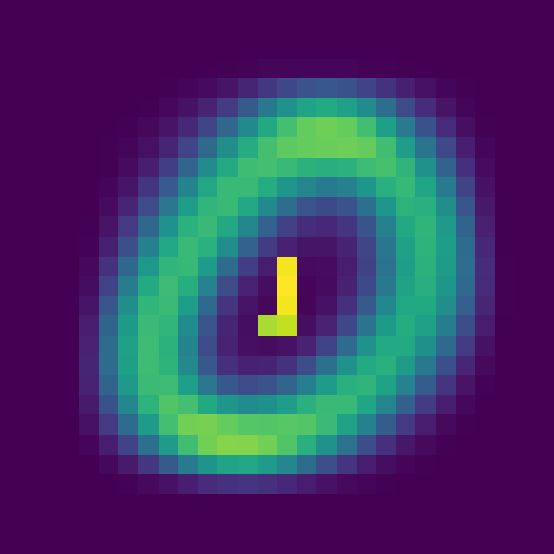}
			\caption{}
		\end{subfigure}
		\begin{subfigure}[b]{0.242\textwidth}
			\includegraphics[width=\textwidth]{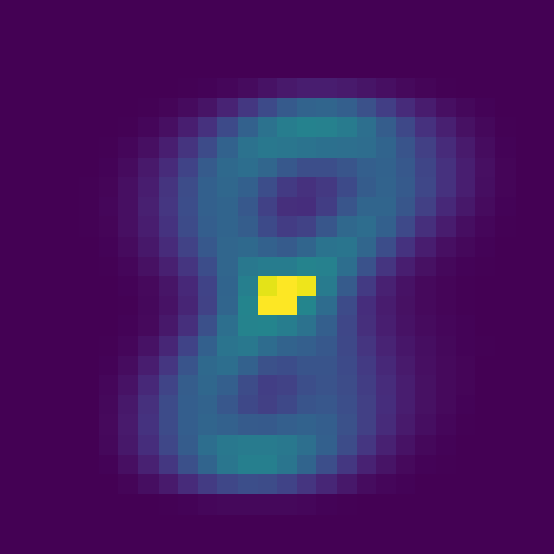}
			\caption{}
		\end{subfigure}
		
		\caption{Distinguish images by difference. We select 5 most important features from the MNIST dataset, these five pixels are located in the center of the images. We can see that these pixels show significant differences between the digits 0 and 8, both in the individual samples ((a) and (b)) and in the class averages ((c) and (d)).}
		\label{Demonstrating ContrastFS on the MNIST dataset.}
	\end{figure}
	However, feature selection methods face several challenges when dealing with high-dimensional data \cite{li2017challenges, song2021variable}.
	One of the main challenges is the efficiency of these methods, as many of them have high computational requirements that make them infeasible for large datasets.
	Another challenge is generality, many methods evaluate features based on predefined models and may not be easily generalized to other models or tasks. 
	Stability is also an issue to consider, many feature selection methods are sensitive to noise or perturbations in the data, and some stochastic model-based methods may not produce consistent results across multiple runs, affecting their interpretability and reliability.
	
	To illustrate the problem conveniently, we categorize existing feature selection methods into two classes: model-based and model-free.
	The former evaluate features by training learning models, while the model-free methods evaluate features based on the intrinsic information in the data.
	In general, model-free methods are more efficient, generic, and stable than model-based approaches, but many of them are not applicable to high-dimensional datasets due to inefficient or ineffective criteria estimation.
	Typically, the criteria could be divided into three groups, similarity-based, information theoretical, and statistics-based \cite{li2018feature}.
	Similarity-based methods try to find features preserve data manifold structure best \cite{zhao2013similarity}, but computing the similarity or distance between samples in high-dimensional feature space could be prohibitively expensive.
	Information theoretical methods measure the relevance and redundancy of features using tools from information theory and most of them could be unified in a conditional likelihood maximization framework\cite{brown2012conditional}, but estimating the joint distribution of features and labels could be challenging and inaccurate for high-dimensional or complex datasets.
	Statistics-based methods are generally fast but the predefined statistical measures may limit the application, e.g., the Pearson's correlation could only detect linear relationships.
	In this paper, we focus on finding an effective and efficient criterion to construct a model-free feature selection method to address these issues.
	
	Identifying such a criterion can be hard, as effectiveness often entails high computational cost.
	Fortunately, in supervised learning, we can circumvent this dilemma by leveraging the label information, whereby the utility of a feature corresponds to the discrepancies it manifests in different classes, and thus we can evaluate the feature.
	This idea is natural, suppose that we need to compare images, we value the regions with the most differences. Similarly, we examine the overall differences between sets of images.
	Therefore, we can intuitively quantify the importance of a pixel based on its values across different images, important features selected based on this criterion can be used intuitively to distinguish images.
	As shown in Figure \ref{Demonstrating ContrastFS on the MNIST dataset.}, 5 central pixels that have the largest differences in value across classes can already distinguish well between the images of 0 and 8 in the MNIST dataset.
	This reveals that, if we can represent each class well as a data point in the original feature space, we can select the salient features based on the discrepancy between them.
	This idea works regardless of the relationship between features and labels, since all the disparity must be grounded on the inherent nature and represented by values.
	Remember the fact that irrespective of the specific form of the data (images, audio, text, etc.), its electronic record is digital, so the difference between two images can always be attributed to the distinctness in values, so that more salient features exhibit greater discrepancy.
	It can also be extended to the general case of classifying samples into different classes with different physical meanings for each feature.
	This is the core idea of our proposed criterion: summarize samples of different classes with a surrogate representation inheriting the original feature set, and evaluate features by the discrepancies between classes.
	
	In this paper, we introduce a new model-free feature selection method that evaluates the importance of features based on statistics inferred from  the data.
	It merits features by measuring how much they vary corresponding to the labels changes (for classification problems, this means the disparity of features among different classes), based on the intuition that good features should change significantly when the output changes.
	Our method computes a surrogate representation of certain class as the statistical individuality contrast to the commonality of the whole dataset, and then evaluates features by computing the contrast that features shown in different classes based on these representations.
	Thus we call our method ContrastFS.
	We also employ these surrogate representations to substitute the original data to study the correlation among features and eliminate redundancy. 
	The most exciting characteristic of our method is that it's much faster than other methods, by a remarkable difference of several orders of magnitude, while attaining best or near the best accuracy performance.
	The main contribution of our work is as follows:
	
	\begin{itemize}
		
		\item We propose a novel feature selection method, called ContrastFS, which constructs surrogate representations to capture the statistical individuality of each class and merits features by quantifying their discrepancies among classes. 
		\item We implement experiments on several real-world datasets and show that our method is fast in computation, clear in meaning, and offers a good trade-off between classification accuracy and running time.
		\item We show that surrogate representations could be used to investigate the correlation between features,  allowing us to improve performance while maintaining the benefits of efficiency.
		\item We show that our method is stable for perturbations and the stability and performance could be enhanced by bootstrap.
		
	\end{itemize}
	
	This paper is organized as follows.
	Section 2 introduces the related work.
	Section 3 formulates the supervised feature selection problem in a probabilistic framework, and then presents our method in detail. 
	Section 4 presents the results of experiments conducted on real-world datasets. 
	Finally, Section 5 concludes this paper.
	
	\section{Related Works}
	\label{releated_works}
	In this section, we briefly review the state of the art in feature selection.
	

	Feature selection is an active research area for decades, there are many methods for feature selection, which can be broadly categorized into three types according to the usage of label information: supervised, semi-supervised, and unsupervised.  \cite{guyon2003introduction,cai2018feature,li2018feature}.
	Supervised feature selection methods leverage the labels of the data to select the most relevant features for classification or regression tasks. 
	Semi-supervised feature selection targets scenarios where labels are rare or where feature relevance and structure are jointly considered, it evaluates feature importance by using the labels of labeled data and the distribution or structure of both labeled and unlabeled data.
	Unsupervised feature selection differs from supervised and semi-supervised feature selection in that it does not use labels to evaluate feature relevance. It measures feature quality based on the intrinsic structure and distribution of the data. Its goal is to find a feature subset that preserves the original data information and reveals the latent patterns or clusters.
	A typical unsupervised feature selection method is UDFS  \cite{yang2011l2}, which selects features that are both discriminative and reconstructive by solving a sparse optimization problem. 
	
	Consider that supervised feature selection can achieve higher accuracy and interpretability for learning task, as it uses the target variable or the output data to select relevant features. In this paper, we focus on the supervised case and review some of the existing methods and challenges.
	
	Supervised feature selection can be divided into three types according to the role of learning algorithms in the feature selection process: filter, wrapper and embedded.
	
	Filter methods evaluate features based on specific optimality criteria without involving any learning algorithm.
	Filter methods are fast, scalable and robust to overfitting, but they may ignore the interactions between the features and the learning algorithm. They typically produce a ranking or a subset of features that can be used as inputs for subsequent learning algorithms.
	
	Wrapper methods use learning algorithms as evaluators to merit features. 
	For example, the HSIC-LASSO \cite{yamada2014highdimensional} uses kernel learning to discover nonlinear feature interactions. 
	Wrapper methods usually achieve good accuracy as the learning model may be able to capture the interaction among features, but the search space is huge, finding the optimal subset of features is NP-hard \cite{kohavi1997wrappers}, making a good approximation is even hard for high-dimensional datasets. 
	
	Embedded methods integrate feature selection into the model learning process. \cite{cai2018feature}. 
	Embedded methods are similar to the wrapper methods in that they all depend on learning models to evaluate features. 
	Embedded methods combine the feature selection problem with model training, usually producing a sparse or optimal model that incorporates the selected features, for which they may be more computationally efficient and less prone to overfitting.
	A typical example in recent years is LassoNet \cite{lemhadri2021lassonet}. 
	The author uses a modified objective function with constraint, integrates training of a neural network with skip connection and feature selection, extends LASSO \cite{tibshirani1996regression} to nonlinear problems, and shows excellent performance in various areas.
	
	Although wrapper and embedded methods usually achieve higher accuracy than filter approaches, models used in the wrapper and embedded methods should be expressive enough to capture the relationship between input and output well, which results in high computational cost and the potential risk of overfitting.
	In addition, the mechanisms of different models may vary significantly, the importance of the feature obtained by these methods is essentially model dependent, which makes the optimal feature subsets obtained under different models may differ. 
	Furthermore, the "optimal" feature subset may vary significantly when a stochastic model is used as the evaluator.
	
	In order to be generic, interpretable, and efficient, we focus on filter methods in this paper.
	
	\subsection{Filter methods}
	Filter methods are a class of feature selection techniques that use specific criteria to evaluate the relevance and importance of features. These methods explore the intrinsic information contained in the data, without relying on any external classifier or predictor. Some common evaluation criteria are based on similarity, information theory, or statistics \cite{li2018feature}.
	
	Many feature selection methods aim to find a feature subset of input which preserve the data similarity best. 
	In supervised feature selection problems, labels in datasets could be leveraged to assess the similarity.
	A representative method such as Fisher Score \cite{duda2000pattern}  tends to find such a feature subset, in which space spanned by these features, the samples from the same class aggregate as close as possible, those from different classes scattered as dispersed as possible. 
	Trace-Ratio \cite{nie2008trace} selects feature subsets via maximizing within classes data similarity while minimizing between classes data similarity. 
	
	Similarity-based methods are usually straightforward and running fast, meanwhile perform well in many tasks. The drawback of these methods is that they rarely take redundancy of selected features into account.
		
	Information-based methods are a type of feature selection methods that use measures of information gain or mutual information between features and the target variable to rank features.
	Although new feature selection methods based on information theory emerge almost every year, most of them can be unified into a conditional information maximization framework\cite{brown2012conditional,li2018feature}.
	JMI \cite{yang1999data} focuses on increasing the complementary information between unselected features and selected features. 
	mRMR \cite{hanchuanpeng2005feature} uses mutual information to select output-relevant features while reducing the feature redundancy. 
	CMIM \cite{fleuret2004fast} selects features to maximize the conditional mutual information with the class to predict  conditional to features selected so far.
	DISR\cite{meyer2008informationtheoretic} uses the normalization techniques and measures the additional information provided by the feature subset with respect to the sum of each individual feature.
	
	Information-theoretic feature selection methods actually provide good physical sense, most of them could consider the redundancy and relevance in feature subsets simultaneously, but the estimation of mutual information without the distribution known in advance could be hard. 
	
	As the natural modeling framework is to consider datasets are drawn from distribution, and tools for sampling and distribution analysis are systematic and abundant in statistics, there are also many methods select feature based on various statistical measures.
	A typical feature selection method in this class is CFS \cite{hall1999feature}, it uses the correlation-based heuristic to merit feature subset, hold the idea that good feature subset should show high correlation with labels while being almost uncorrelated with each other.

	Statistics-based feature selection methods are often simple, intuitive and computational efficiently, but most of them rely on certain pre-set statistical measures, which limit the applicability of methods.
	
	\section{Methods}
	We first formulate the feature selection problem in Section \ref{problem_formulation}, then we introduce our solution in Section \ref{our_method}.
	Our method is presented in two parts: class discrepancy estimation and feature selection.
	In the former, we show how to construct surrogate representations based on low-order sample moments and measure feature discrepancies.
	In the latter, a discriminative subset of features is selected according to the discrepancies they show among different classes.
	We show how to tackle the redundancy problem with respect to the proposed algorithm in Section \ref{reduce_redundancy}.
	Then we discuss the statistical properties of our method in Section \ref{Statistical_Properties}.	
	
	\subsection{Problem formulation}
	\label{problem_formulation}
	The goal of feature selection problem is to find a subset $\mathcal{T}^{\star}$  from original feature set $\mathcal{S}=\{{f_1},\cdots,{f_d}\}$ which has maximal utility $U$, this problem often constrained by the cardionality of $\mathcal{T}$ to be specific number $m$.
	\begin{equation}
		\mathcal{T}^{\star} =\arg \max _{\mathcal{T} \subseteq \mathcal{S}} U(\mathcal{T}),\text { s.t. }|\mathcal{T}| =m,\quad m<d \label{original_fs_problem}
	\end{equation}
	where $|\cdot|$ is the cardinality of a set.
	
	
	Consider features as a high-dimensional random variable with probability distribution $p(\mathbf{x})$ over a $d$-dimensional space, samples are drawn from it.
	We can reformulate the goal of original feature selection problem\ref{original_fs_problem} in probabilistic framework: find a subset $\mathcal{T}^{\star}$ of features from original feature set $\mathcal{S}$ with specified size $m, m<d$, which maximizes some utility $U(\cdot)$ in expected sense.
	
	\begin{equation}
		\mathcal{T}^{\star} =\arg \max _{\mathcal{T} \subseteq \mathcal{S}} E_p[U(\mathcal{T})],\text { s.t. }|\mathcal{T}|=m ,\quad m<d \label{prob_fs_problem}
	\end{equation}
	
	
	In practice, we do not know the $p(\mathbf{x})$ exactly, all we know is a data matrix $X\in R^{d\times n}$ consists of $n$ samples drawn from $p(\mathbf{x})$.
	We need to reformulate the feature selection problem\ref{prob_fs_problem} to an empirical form. 
	Here, the goal becomes choosing a sub-matrix $X_{\mathcal{T}^\star}\in R^{m\times n}$ from data matrix $X$ to achieve optimality under certain criteria $F$:
	
	\begin{equation}
		\mathcal{T}^*=\arg \max _{\mathcal{T} \subseteq \mathcal{S}} F(X_\mathcal{T}), \text { s.t. }|\mathcal{T}|=m ,\quad m<d \label{emp_fs_problem}
	\end{equation}
	
	where $F(X_\mathcal{T})=\hat{U}(\mathcal{T})$, and $\hat{U}(\mathcal{T})$ is the estimation of utility of feature subset $\mathcal{T}$.
	
	There are two main difficulty to solve this problem, one is how to evaluate the utility of feature subset which is consistent to consequent analysis, the other is that this combinatorial problem is NP-hard.
	We will introduce our solution to this problem in next section.
	
	\subsection{Our Proposal}
	\label{our_method}

	In this section, we propose our solution to address the problem mentioned above by leveraging label information and the heuristic top-$k$ strategy.
	
	Under the supervised classification setting, we can leverage the label information to evaluate the features, as the relationship between the input variable $\mathbf{x}$ and the label variable $\mathbf{y}$ is the core, the utility of feature can be represented as discriminative power.
	Assuming that the samples in the data matrix $\mathbf{X}$ correspond to $C$ classes, the feature selection problem can be reformulated as follows:

	\begin{equation}
		\mathcal{T}^{\star} =\arg \max _{\mathcal{T} \subseteq \mathcal{S}} \sum_{i=1}^C\sum_{j=1}^C F_d(X_{\mathcal{T}}^i,X_{\mathcal{T}}^j),\text { s.t. }|\mathcal{T}|=m \label{prob_dist_problem}
	\end{equation}
	
	where $F_d(\cdot,\cdot)$ is the discriminative function which estimates the discriminative power of features based on data from different classes corresponding to features, $X_{\mathcal{T}}^i,X_{\mathcal{T}}^j$ are the samples of the $i$-th and $j$-th class corresponding to feature subset $\mathcal{T}$ respectively.
	
	The choice of the discriminant function is now the core of a feature selection method, more expressive choices for $F_d(\cdot,\cdot)$ will allow for better performance in subsequent data analysis but also require higher computational cost. 
	A natural intuition is that good features vary significantly across different classes, in other words, they contribute to distinguish samples from different classes.
	Otherwise, if the values of a feature nearly the same when the label changes, the feature could not provide useful information for prediction.
	This links the discriminative power of features among different classes to the discrepancies, so we can choose the discrepancy measure $D(\cdot,\cdot)$ as the discriminant function. 
	That is, using the discrepancies of features shown across classes as the criterion to evaluate the importance of features.
	\begin{equation}
	\mathcal{T}^{\star} =\arg \max _{\mathcal{T} \subseteq \mathcal{S}} \sum_{i=1}^C\sum_{j=1}^C D(X_{\mathcal{T}}^i,X_{\mathcal{T}}^j),\text { s.t. }|\mathcal{T}|=m \label{discrepancy_problem}
	\end{equation}

	We can further simplify the problem heuristically, a common strategy is to evaluate features individually and select features greedily. 
	Now the selected feature subset $\mathcal{T}^\star$ consists of the top-$m$ features ${f}_t,t=1,\cdots,m$ with highest magnitude of discrepancy.
	\begin{equation}
		\mathcal{T}^{\star} =\arg \max _{\mathcal{T} \subseteq \mathcal{S}}\sum_{t=1}^m \sum_{i=1}^C\sum_{j=1}^C D(X_{{f_t}}^i,X_{{f_t}}^j),\text { s.t. }|\mathcal{T}|=m \label{top_m_problem}
	\end{equation}

	What we concern now is how to measure the discrepancy to evaluate the utility of features.
	We are inspired by the fact that we can distinguish samples or groups of samples by features with large differences over the mean of the data \ref{Demonstrating ContrastFS on the MNIST dataset.}. Thus, it  is feasible to use a surrogate representation to summarize samples of each class, and evaluate the importance of features by the discrepancy between classes.
	This leads to the core of our method, to construct surrogate representations to summarize the statistical behavior of samples in each class, so that the magnitude of the discrepancy across classes represents the discriminative power of features straightforwardly.
\begin{figure}[]
	\centering
	
	\begin{subfigure}[b]{0.24\textwidth}
		\includegraphics[width=\textwidth,trim={0 0 0 1cm},clip]{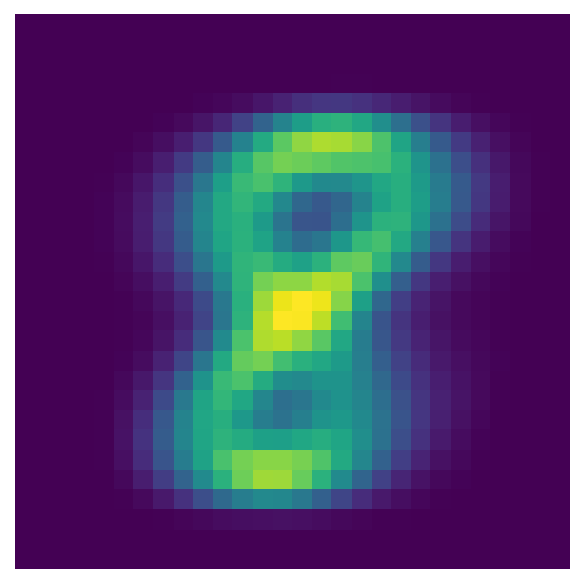}
		\caption{}
		\label{mean_of_data}    
	\end{subfigure}
	\begin{subfigure}[b]{0.24\textwidth}
		\includegraphics[width=\textwidth,trim={0 0 0 1cm},clip]{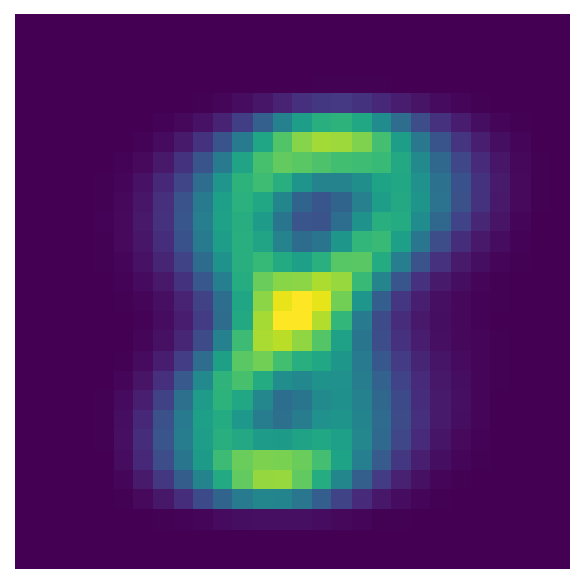}
		\caption{}
		\label{l2barycenter}    
	\end{subfigure}
	\begin{subfigure}[b]{0.24\textwidth}
		\includegraphics[width=\textwidth,trim={0 0 0 1cm},clip]{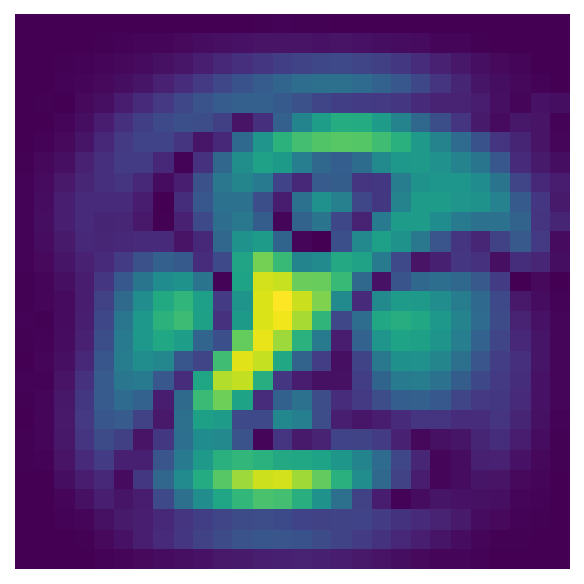}
		\caption{}
		\label{standardization}    
	\end{subfigure}
	\begin{subfigure}[b]{0.24\textwidth}
		\includegraphics[width=\textwidth,trim={0 0 0 1cm},clip]{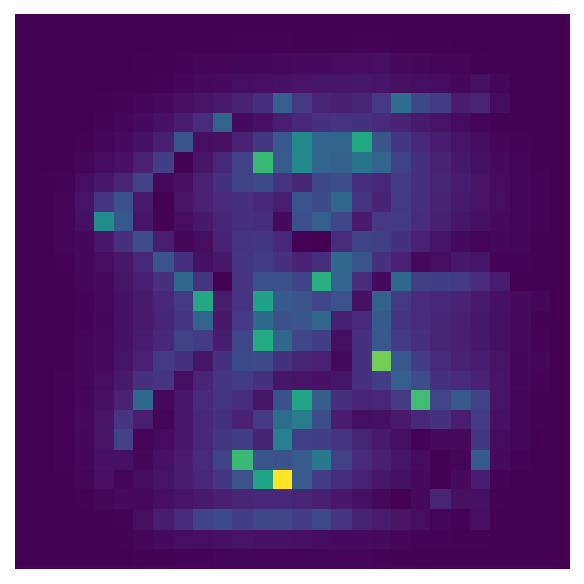}
		\caption{ }
		\label{Surrogate_representation}    
	\end{subfigure}	
	\caption{Representation of class 8 of the MNIST dataset. (a) shows the average of the samples belonging to digit 8. (b) shows the $l_2$ barycenter of the samples. (c) shows the normalized plot, which removes the mean and scales by the standard deviation. (d) shows the surrogate representation, which captures the deviation from the commonality of the original data by a dimensionless quantity.}
		\label{Dimensionless representation of MNIST dataset.}
	\end{figure}
	
	\subsubsection{Construct Surrogate Representation}

	We now discuss how to construct surrogate representations of each class for datasets from different domains.
	To estimate the discrepancies between classes properly, we need to revisit the modeling of datasets and take normalization, due to noise and heterogeneity contained in datasets are commonly seen in practice.
	Consider a dataset consists of $n$ samples $\{\mathbf{x_i},y_i\}(i=1,\dots,n)$, each sample  consisting of observation about $d$-dimensional input variables $\mathbf{x}$ and one output variable $\mathbf{y}$. 
	$$\{ ( \mathbf{x _ { i }} , y _ { i } ) | x _ { i }  \in R^d ,  y _ { i } \in R , i=1,\cdots , n \}$$
	
	We assume samples are independent and identically distributed $(i.i.d)$ paired drawn from a joint distribution with density $p(\mathbf{x},\mathbf{y})$, then we separate those pairs and denote original data by $\mathbf{X},\mathbf{Y}$:
	\begin{equation}
		\begin{split}
			\mathbf{X} & = [ \mathbf{x }_ { 1 } , \cdots , \mathbf{x} _ { d } ]^T \in R^{d\times n} \\
			\mathbf{Y} & = [ y _ { 1 } , \cdots , y _ { n } ] ^T \in R^n
		\end{split}
	\end{equation}
	where $\mathbf{x} _ { k } = [x_{k,1}, \dots, x_{k,n} ]^T \in R^n$ is a vector consists of values corresponding to the $k$-th feature.  
	
	A number of methods are potential candidates for constructing surrogate representation.
	A natural idea is to represent each class by the mean of the data, but it may not work if the means of the samples are almost unchanged across classes, and uniform weighting of features is often inappropriate.
	The barycenter is also a typical candidate for summarizing samples, for the ability to account for the relationship between samples.
	However, weighting features based on their magnitude in datasets may lead to ignorance of informative ones.
	As is shown in Figure \ref{Dimension-less representation of MNIST dataset.}, both the mean of data and $l_2$ barycenter for class could be distinguished directly, but the edge of the digit shown in figure obtained by the $l_2$ barycenter is smooth, implying the omission of informative features even though they may be very distinctive across classes.
	In addition, since the features in datasets often have different physical meanings and value ranges, a unified dimension or dimensionless quantity is required for comparison when performing feature selection.
	
	A common method is to perform standardization.
	The rationale of standardization lies in two aspects: first, it provides a dimensionless quantity that can be intuitively used for comparison; second, as a random variable, the feature will asymptotically follow a normal distribution after standardization under mild assumptions.
	In this way, we can summarize each class like normalization based on the empirical expectation, the standard deviation of each class and the whole dataset, respectively:
	\begin{equation}{\phi}_t^k = \frac {{\mu_t^k}-{\mu_t}}{{\sigma_t}}, \quad t\in\{1,\dots,{d}\} \quad k\in\{1,\dots,C\} 
	\label{surrogate_representation_normalization}
	\end{equation}
	where ${\phi_t^k}\in R$ and ${\mu_t^k}$ are the surrogate representation constructed by standardization and the unbiased estimation of mean of $k$-th class corresponding to feature ${f_{t}}$ respectively.
	${\mu}_t, {\sigma}_t$ are the unbiased estimation of expectation and standard deviation corresponding to feature ${f_t}$ on the whole dataset respectively.
	
	However, it does not satisfy our need: to describe the distinctiveness of features in different classes.
	Features usually represent different distribution properties across classes, while the formula \ref{surrogate_representation_normalization} only reflects the deviation of the mean of the feature on samples corresponding to a particular class from its mean on the whole dataset.
	In practice, it's more reasonable to model the feature in different classes as different random variables.
	Note that it is still reasonable to describe the statistical behavior of a class conditional distribution based on the low-order sample moments without prior knowledge and to construct a dimensionless quantity for comparison, so we modify the formula \ref{surrogate_representation_normalization} to achieve our goal.
	We scale all features to the same range and compute low-order sample moments based dimensionless quantities as surrogate representation of each class for the following analysis.
	We consider the mean and standard deviation of the features in each class, and scale them to eliminate the influence of dimension and subclass.
	We also take the coefficient of variation $C_v$ into account, to weight the feature to flexibly meet the challenges of different domains.
\begin{equation}
	{Z}_t^k = Cv_t^k \frac {{\mu_t^k}-{\mu_t}}{{\sigma_t^k}-{\bar{\sigma_t}}}, \quad i\in\{1,\cdots,{d}\} \quad k\in\{1,\cdots,{C}\} 
	\label{surrogate_representation}
\end{equation}
where the $\bar{\sigma_t}$ is the mean of ${\sigma_t}^k$.	The $C_v$ should be set with respect to certain case, the $C_v$ is recommended to be set as 1 to retain the distributional structure of original data if without prior knowledge of data.
\begin{equation}
		Cv_t^k = \begin{cases}
		{ \frac { \sigma_t^k } { \mu_t^k } },\quad highlight ~ relative~ standard~ deviation\\ 
		{ \frac { \mu_t^k} { \sigma_t^k}, \quad highlight~ stable~ variables}\\
		{1},\quad others
		\end{cases}
\label{coef_of_variation}
\end{equation}

	The representation $\mathbf{Z}^k$ is rather the individuality of class $k$ shown by the features rather than the average of the samples, as it reflects the difference of the statistical structure of class $k$ compared to the whole dataset, as shown in Figure \ref{Dimensionless representation of the MNIST dataset.}.

	The highlight of this procedure is that it summarizes the samples of each class to a new data point in the original feature space, so that we can get rid of the scale of the datasets in subsequent analysis and study features fairly.
	
	\subsubsection{Evaluate Features}

	In the supervised setting, the discriminative power of certain feature is the distributional discrepancies it behaves across classes, that is, the disparity of the empirical distributions across different classes.
	Now we show how to find the informative features.
	Given surrogate representations of each class, we compute the average discrepancies in a pairwise manner across different classes as the importance score $I(f_t)$ of the feature ${f_t}$.
	\begin{equation}
		I({f_t})= \frac{1}{C(C-1)}\sum_{i}\sum_{j,j\neq i}
		|Cv_t^k \frac {{\mu_t^i}-{\mu_t}}{{\sigma_t^i} - \bar{\sigma_t}}
		-Cv_t^k \frac {{\mu_t^j}-{\mu_t}}{{\sigma_t^j} - \bar{\sigma_t}}| ,
		\quad i,j\in\{1,\cdots,{C}\}
	\label{xor_operation}
	\end{equation}
	It's worth noting that all features of the input could be evaluated simultaneously in a vectorized manner.
	
	Based on this evaluation, we rank all features of the input and select the top-$m$ to be the best feature subset with size $m$,  this is our proposed ContrastFS: select important features based on the contrast of values corresponding to them between classes.
	The difference between surrogate representations reveals the basis to distinguish samples from different classes, so that the selected features  have significant discriminative power.
	In figure \ref{Demonstrating ContrastFS on the MNIST dataset.}, we select the most 5 discriminative pixels from 784 pixels of images in the MNIST dataset, we notice that we can distinguish images directly based on these pixels.
	
	\subsection{Reduce the Redundancy}
	\label{reduce_redundancy}
	Since we evaluate features individually, the feature subset selected by the top-m strategy may has redundancy.
	If prior knowledge is available, manual processing by domain expert may be the best solution to reduce the redundancy, in which case ContrastFS acts as a griddle to narrow the range of candidate features. 
	If no prior knowledge is available, as is often the case in practice,	we need to find an appropriate measure to detect and reduce the redundancy.
	In this section,  we show how to explore the redundancy in the feature set using surrogate representations, and thus to reduce the redundancy if necessary.
	
	With surrogate representations at hand, we can study the similarity between them by an appropriate similarity measure to explore redundancy in features.
	Since we focus on the discrepancies of features shown across classes, an intuitive idea is that if the discrepancies correspond to feature ${f_i}$ that are similar to their counterparts of ${f_j}, j\neq i$ across classes, these two features may be replicated, we can prune the one with smaller importance score computed in \ref{xor_operation}.
	\begin{equation}
		Redundancy(\{f_i, f_j\}) = Similarity(\mathbf{D}_{{f_i}}(\cdot,\cdot),\mathbf{D}_{{f_j}}(\cdot,\cdot))
	\label{reduce_redundancy_1}
	\end{equation}
	where $\mathbf{D}_{{f_i}}, \mathbf{D}_{{f_j}}$ are vectors consisting of discrepancies shown by feature $f_i, f_j$ across classes respectively.
	It's worth noting that in this formulation, the measure of redundancy among features does not involve the original dataset, which reduces the computational burden imposed by the scale of the original dataset size.
	The extension of this method to investigate the redundancy among multiple features is contingent on the choice of the similarity measure.
	
	To measure and reduce the redundancy, we provide a heuristic method by means of Pearson's correlation, denoted by $Cor(\cdot,\cdot)$. 
	We compute the correlation between the discrepancies of selected features in a pairwise manner as the measure of redundancy.
	Then we can formulate the \ref{reduce_redundancy_1} as follows:
	\begin{equation}
		Redundancy(\mathcal{T}) = \sum_{i=1}^{m}\sum_{j=1}^{i}Cor(D_{{f_i}}(\cdot,\cdot),D_{{f_j}}(\cdot,\cdot)),\quad f_i \in \mathcal{T}
	\label{reduce_redundancy_2}
	\end{equation}
	Computing the correlation in a pairwise manner will give us a $m$ dimensional matrix, we heuristically take the average of each row as the measure of the redundancy brought by the corresponding feature, then prune the most redundant ones to reduce the redundancy of $\mathcal{T}$.

	\subsection{Statistical Properties of Our Methods}
	\label{Statistical_Properties}
	\textbf{Convergence}:
	In our opinion, the most important property of the feature selection criterion is the convergence, as the statement proposed in  \cite{song2012feature}, it should be concentrated with respect to the underlying measure to ensure the effectiveness when faced new samples.
	The evaluation criterion in our method is based on the unbiased estimation of the low-order moments of the data distribution, with the common i.i.d. assumption made on the data generation distribution, the law of large numbers guarantees its convergence\cite{wasserman2010all}. 
	Moreover, if prior knowledge of the data generating distribution is available, the estimation of the low-order moments may converge faster than in the ordinary case, which means that fewer samples are required to achieve stable feature evaluation.
	
	\textbf{Stability}:
	ContrastFS is on the basis of inferring the low-order moments based on the samples of certain class, it's stable and accurate for perturbations, especially when the size of the dataset is large.
	Cooperation with additional statistical techniques can bring benefits to certain task, for example, we can utilize bootstrap to improve the performance and stability of our method.
	
	\textbf{Computation}:
	The most salient feature of our method is its computational efficiency and versatility.
	Our method relies on estimating the low-order moments of the samples, and all the operations involved are elementary ones, which makes our method computationally inexpensive and applicable on various devices.
	Moreover, our method evaluates features individually, thus vectorized or parallel computation can further speed up the process.
	
	\section{Experiments}

	In this section, we conduct experiments \footnote{github URL will come soon.}  to evaluate the effectiveness of our proposed method on real-world datasets and compare it with existing popular feature selection methods. 
	\subsection{Data Sets}
	
	We use benchmark datasets collected from different domains, including image data, speech data, sensor data, and biomedical data, all of which have been used in the literature for benchmarking feature selection methods \cite{balin2019concrete, lemhadri2021lassonet}.
	
	These datasets can be divided into two classes depending on their background and tasks: pattern detection (COIL-20, MNIST, Fashion-MNIST, and nMNIST-AWGN) and tabular datasets (Activity, ISOLET, and MICE). Each sample from the former dataset is an image, while the samples in the latter dataset are structural elements. We experiment on these datasets to compare the effectiveness and versatility of our method with existing approaches. 
	Table \ref{details_of_datasets} summarizes the features of the datasets used in our experiments.\footnote{We use the same preprocessing step as \cite{lemhadri2021lassonet} on COIL-20 dataset, resizing the images to 20*20 images, thus the total number of features becomes 400.}.
	
	\begin{itemize}
	\item COIL-20: The COIL-20 dataset is a collection of 1,440 grayscale images of 20 objects, each of which was captured in 72 different poses. This dataset has been widely used in computer vision research for object recognition and classification tasks.
	
	\item MNIST: The MNIST dataset is a standard benchmark for computer vision and machine learning, consisting of 60,000 handwritten digits for training and 10,000 for testing. It has been extensively used in research on deep learning, image recognition, and pattern recognition.
	
	\item Fashion-MNIST: The Fashion-MNIST dataset is a more challenging and realistic alternative to the original MNIST dataset, consisting of 70,000 images of fashion items. This dataset has been used to evaluate the performance of various machine learning algorithms for image classification tasks.
	
	\item nMNIST-AWGN: The nMNIST-AWGN dataset is an extension of the MNIST dataset created by adding white Gaussian noise to the original images to simulate noisy environments. This dataset has been used to evaluate the robustness of machine learning algorithms to noise in image classification tasks.
	
	\item Human Activity Recognition (Activity): The Human Activity Recognition (Activity) dataset contains numerical items collected by sensors in smartphones, each item corresponding to one of six poses of the user. This dataset has been used to evaluate the performance of various machine learning algorithms for activity recognition tasks.
	
	\item ISOLET: The ISOLET dataset is a collection of spoken letters of the English alphabet, where the cardinality of each sample is a feature extracted from voice data. This dataset has been used to evaluate the performance of various machine learning algorithms for speech recognition tasks.
	
	\item MICE: The MICE dataset is a collection of gene expression data of 77 proteins from 1080 mice with different genetic backgrounds and treatments. This dataset has been used to evaluate the performance of various machine learning algorithms for gene expression analysis.
	\end{itemize}

	\begin{table}[!h]
	\centering
	\begin{adjustbox}{max width=\textwidth,keepaspectratio} 
	\begin{tabular}{|c|c|c|c|c|c|c|c|}
		\hline
		Dataset & Activity & COIL-20 & ISOLET & MICE & MNIST & Fashion-MNIST & nMNIST-AWGN \\ \hline
		\#(Features) & 561 & 400 & 617 & 77 & 784 & 784 & 784 \\ \hline
		\#(Samples) & 5744 & 1440 & 7797 & 1080 & 70000 & 70000 & 70000 \\ \hline
		\#(Classes) & 6 & 20 & 26 & 8 & 10 & 10 & 10 \\ \hline
	\end{tabular}
	\end{adjustbox}
	\caption{The Details of Benchmark Datasets}
	\label{details_of_datasets}
\end{table}

	We randomly split each dataset into training and test sets, to compare fairly, the training and test sets are the same for all methods by dividing according to the same random state. 
	We split the COIL-20, MICE and ISOLET datasets into training and test sets with a 70-30 ratio, as they have relatively few samples compared to the number of classes. 
	For the MNIST, Fashion-MNIST and nMNIST-AWGN datasets, which have a sufficient number of samples, we randomly split them into training and test sets with a ratio of 20-80 to avoid memory overflow on our PC\footnote{The memory requirement of Trace-Ratio and UDFS exceeds the physical memory of our PC if we follow the 70-30 split strategy.}.

	\subsection{Methodology}
	We compare our method with several filter feature selection methods mentioned in the Related Works Section, including JMI \cite{yang1999data}, CFS \cite{hall1999feature}, Fisher Score \cite{duda2000pattern},  Trace-ratio \cite{nie2008trace}, mRMR \cite{hanchuanpeng2005feature}, CMIM \cite{fleuret2004fast},  DISR \cite{meyer2008informationtheoretic}, UDFS \cite{yang2011l2}.
	We also compare popular wrapper and embedded approaches, such as HSIC-Lasso \cite{yamada2014highdimensional},  LassoNet \cite{lemhadri2021lassonet}. 
	Where available, we made use of the scikit-feature implementation \cite{li2018feature} of each method. 
	We implement HSIC-Lasso via package pyHSIC \cite{yamada2014highdimensional} \cite{climente-gonzalez2019block}, and download the source code from the official repository\footnote{https://github.com/lasso-net/lassonet} to implement LassoNet. 
	We set the hyper-parameters of compared methods at default.
	
	We feed the feature sets selected by different methods as input into independent classifiers to explore the effectiveness of each methods for fairness.
	We run each feature selection method in comparison to select a varying number of features, and measure the accuracy obtained by downstream classifiers when these features are fed in as the metric to quantify the performance of feature selection methods.
	We evaluate the efficiency of each method by measuring the time required to evaluate all input features and to train the corresponding classifier.
	We perform a direct comparison of the running time of each method, and examine the influence of the efficiency of feature selection methods on the overall machine learning pipeline.
	
	We report the average performance and time consumption of feature selection methods from 10 runs on different training sets, to be fair, we randomly split the dataset and ensure that all methods work on the same data in each run.
	We implement the experiments in Python on a desktop with an Intel Core i7-12700KF CPU @4.6GHz and 32GB physical RAM.

	\begin{figure}[]
	\centering
	\begin{subfigure}[b]{0.48\textwidth}
		\includegraphics[width=\textwidth]{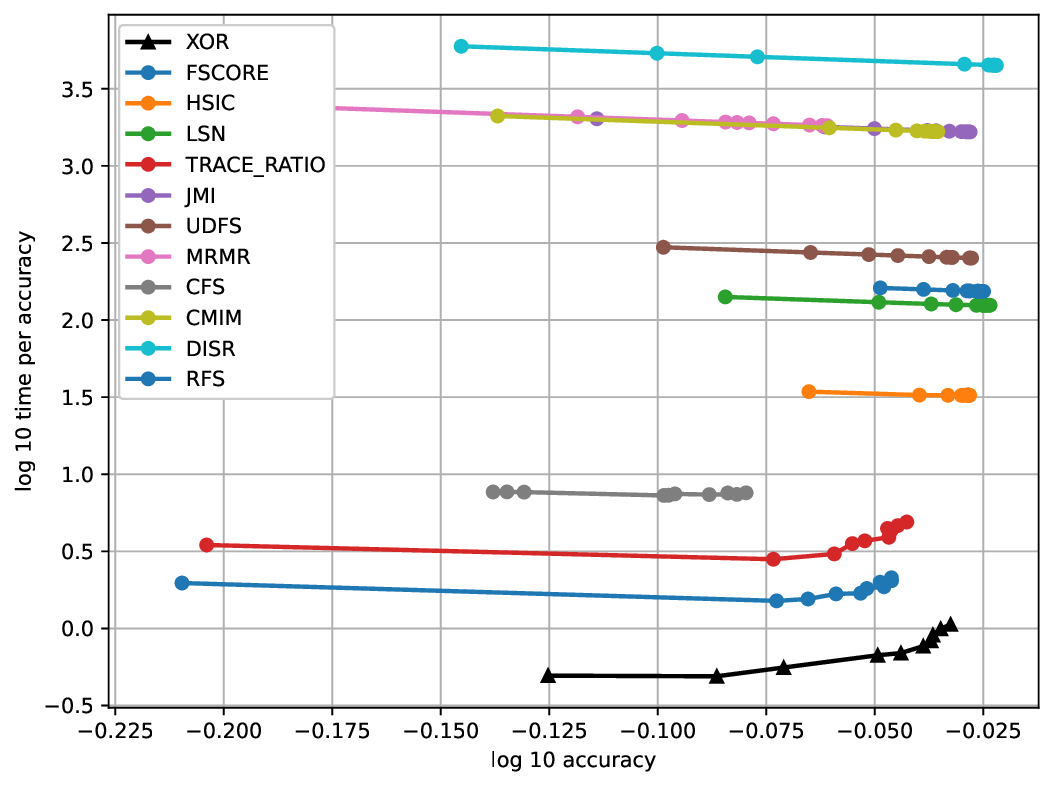}
		\caption{Activity}
		\label{Activity_log10_time_per_acc_xgb}    
	\end{subfigure}
	\begin{subfigure}[b]{0.48\textwidth}
		\includegraphics[width=\textwidth]{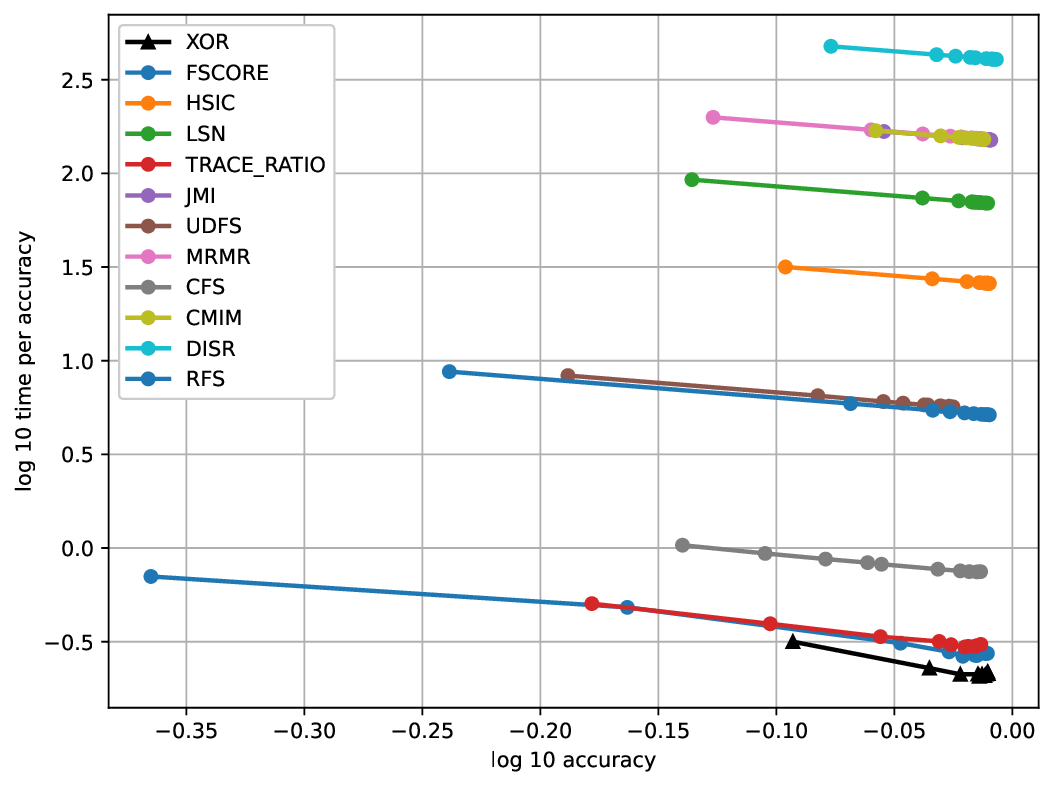}
		\caption{COIL-20}
		\label{COIL-20_log10_time_per_acc_xgb}    
	\end{subfigure}
	\begin{subfigure}[b]{0.48\textwidth}
		\includegraphics[width=\textwidth]{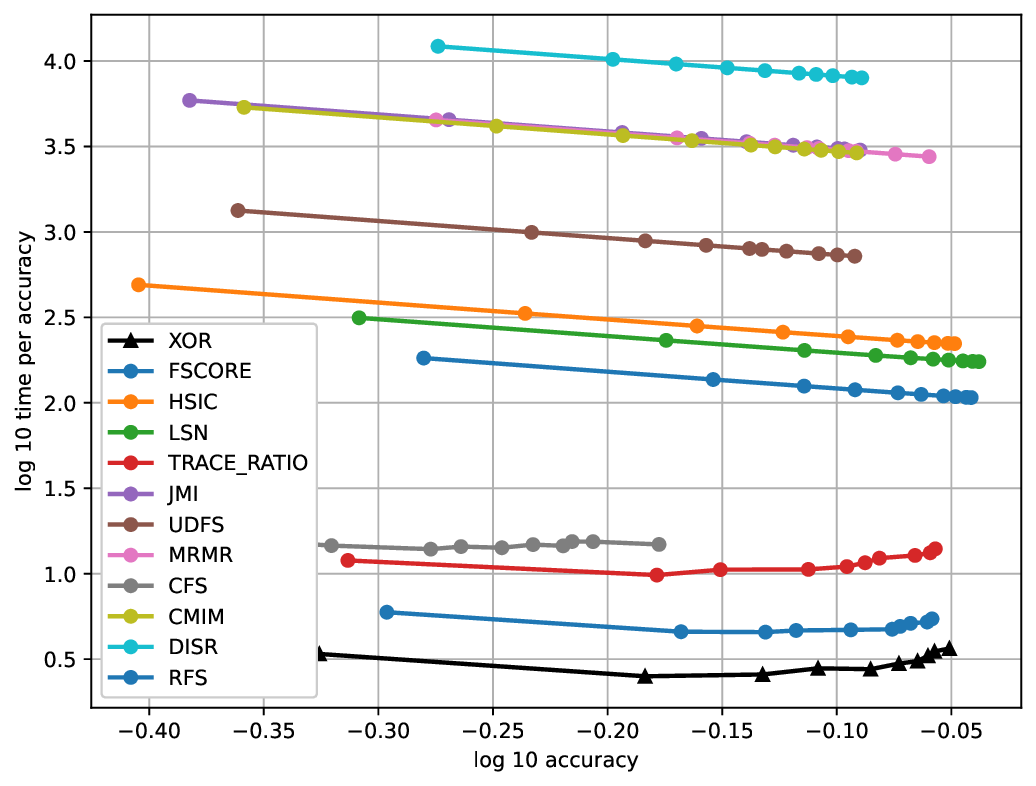}
		\caption{ISOLET}
		\label{ISOLET_log10_time_per_acc_xgb}    
	\end{subfigure}
	\begin{subfigure}[b]{0.48\textwidth}
		\includegraphics[width=\textwidth]{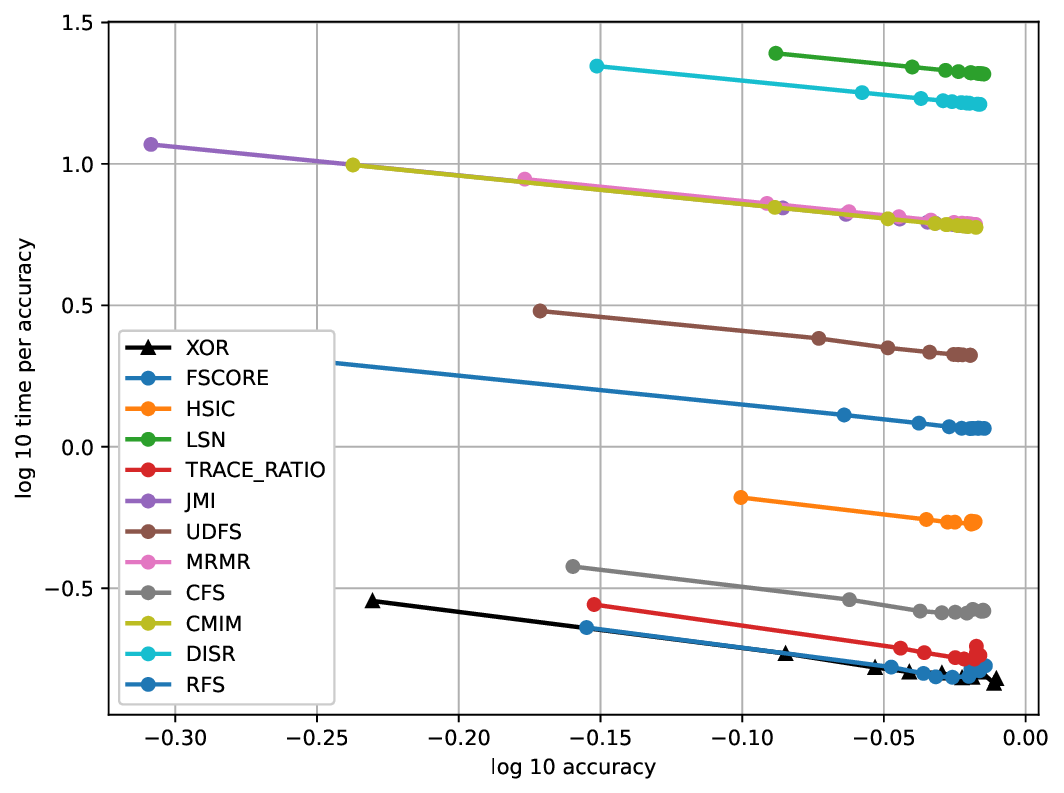}
		\caption{MICE}
		\label{MICE_log10_time_per_acc_xgb}    
	\end{subfigure}
	\begin{subfigure}[b]{0.48\textwidth}
		\includegraphics[width=\textwidth]{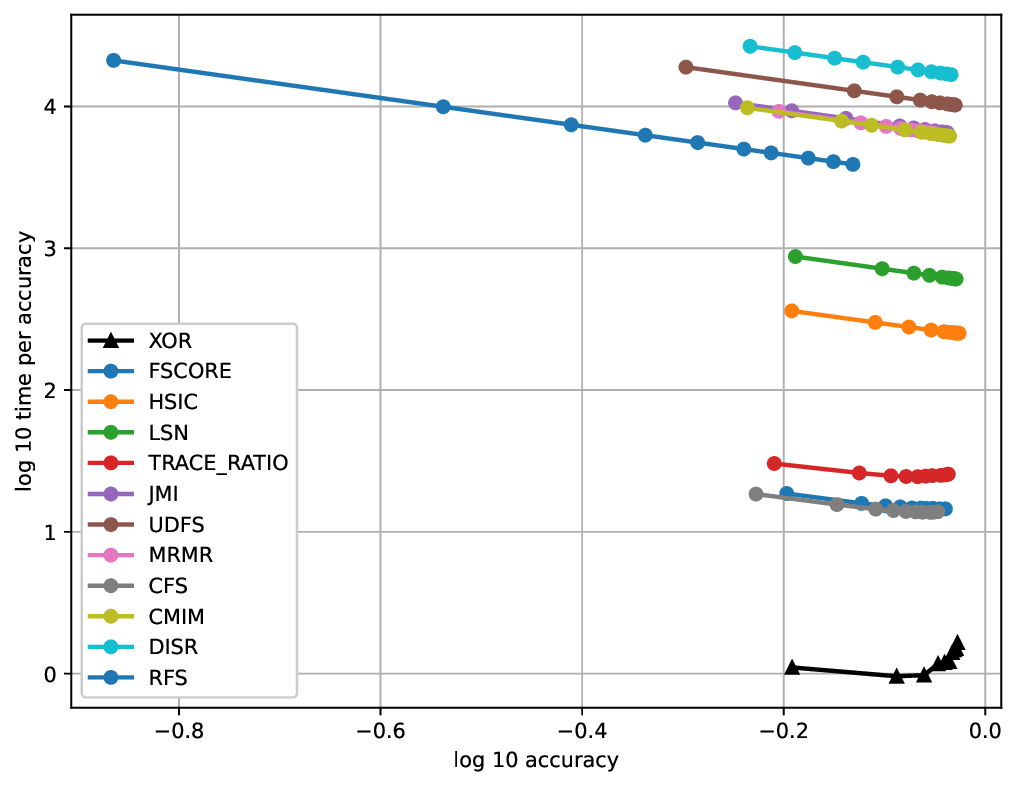}
		\caption{MNIST}
		\label{MNIST_log10_time_per_acc_xgb}    
	\end{subfigure}
	\begin{subfigure}[b]{0.48\textwidth}
		\includegraphics[width=\textwidth]{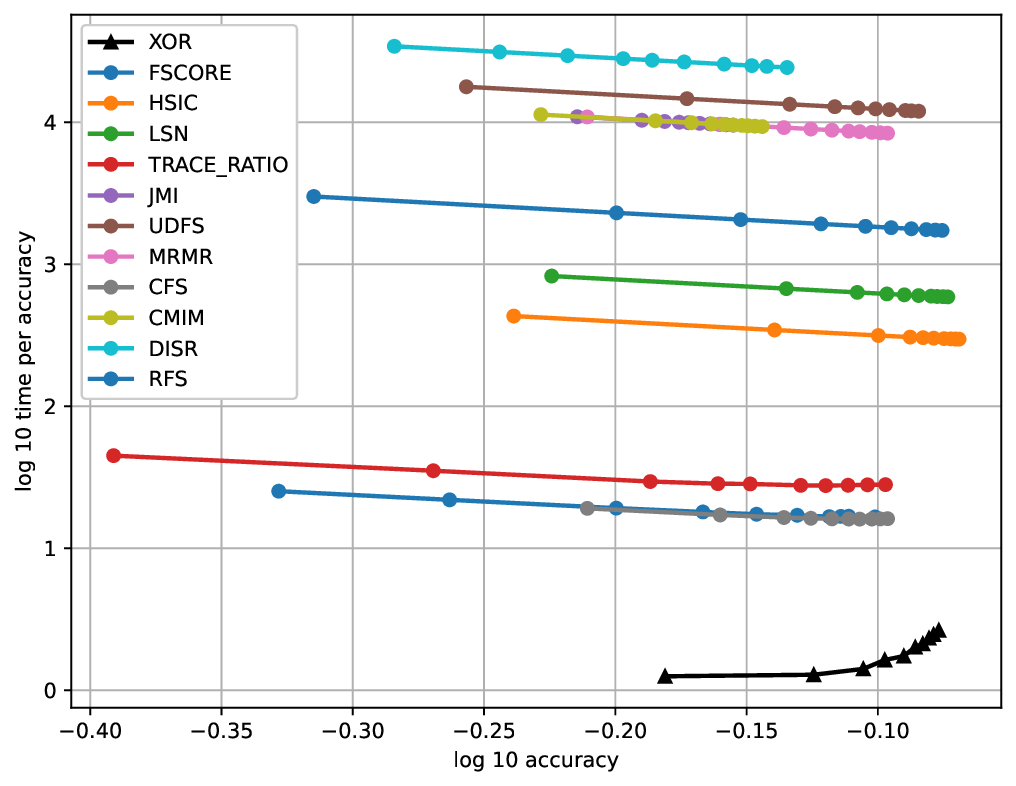}
		\caption{MNIST-Fashion}
		\label{MNIST-Fashion_log10__time_per_acc_xgb}    
	\end{subfigure}
	\caption{Logarithmic unit time cost vs. accuracy.}
	\label{log_unit_gain_vs_acc}
\end{figure}

	\subsection{CPU Time}

	The most interesting characteristic of our method is that it runs very fast. 
	While the feature selection procedure could reduce the computational cost in learning tasks, the feature selection method may take too much time to find out informative feature subsets, especially when dealing with high-dimensional data or complex models. 
	In practice, feature selection is usually performed as part of the overall task pipeline, and as the time cost of feature selection $T_{fs.}$ increases, so does the total time consumption $T_{tot.}$.
	Thus, we test the efficiency of the feature selection approaches mentioned in Section \ref{releated_works}, the results show the huge superiority of our method in terms of running time.

\begin{equation}
T_{tot.} = T_{fs.} +T_{train}
\label{machine_learnig_time_cost}
\end{equation}

\begin{table}[!ht]
	\centering
	\begin{adjustbox}{max width=\textwidth,keepaspectratio}
	\begin{tabular}{|c|r|r|r|r|r|r|r|}
		\hline
		  & Activity & COIL-20 & ISOLET & Mice & MNIST & Fashion-MNIST & Mean \\ \hline
		ContrastFS & 1 & 1 & 1 & 1 & 1 & 1 & 1 \\ \hline
		FisherScore & 53 & 18 & 69 & 54 & 223 & 217 & 168 \\ \hline
		CFS & 301 & 158 & 199 & 250 & 204 & 216 & 218 \\ \hline
		Trace-Ratio & 77 & 25 & 103 & 76 & 332 & 319 & 249 \\ \hline
		HSIC-Lasso & 1636 & 7894 & 7048 & 766 & 4630 & 4818 & 4836 \\ \hline
		LassoNet & 6597 & 21217 & 5625 & 39871 & 11384 & 9564 & 9462 \\ \hline
		CMIM & 87788 & 46418 & 87175 & 11197 & 114386 & 130389 & 110063 \\ \hline
		mRMR & 90488 & 46713 & 89196 & 11453 & 116050 & 130711 & 111418 \\ \hline
		JMI & 88764 & 46362 & 90532 & 11169 & 120521 & 129950 & 112677 \\ \hline
		UDFS & 13480 & 1606 & 21532 & 3718 & 192319 & 191790 & 135621 \\ \hline
		DISR & 243942 & 125627 & 241210 & 31037 & 312396 & 347178 & 298554 \\ \hline
	\end{tabular}
\end{adjustbox}
	\caption{Relative time consuming of feature selection.}
	\label{Relative_time_consuming}
\end{table}

	We compared the running time of each feature selection method for evaluating all features on these datasets.
	For fairness, we randomly split these datasets to ensure that all methods processed the same data, and we run 10 trials and averaged the results.
	Since the methods used for comparison with ContrastFS are not of the same type, we focus here on the time it takes for each method to evaluate all the features.
	As shown in the table \ref{Relative_time_consuming}, ContrastFS has several orders of magnitude superiority in running time compared to the other methods.
	These results show that our method could be scalable to large scale datasets.
	It's worth noting that our method could be further accelerated by running it in parallel when faced with large scale datasets. 

\begin{table}[]
	\centering
	\begin{adjustbox}{max width=\textwidth,keepaspectratio}
		\begin{tabular}{|c|c|c|c|c|c|c|c|}
			\hline
			  & Activity & COIL-20 & ISOLET & Mice & MNIST & Fashion-MNIST & Mean \\ \hline
			ContrastFS & 0.018 & 0.003 & 0.027 & 0.001 & 0.050 & 0.051 & 0.021 \\ \hline
			FisherScore & 0.928 & 0.060 & 1.876 & 0.027 & 11.118 & 11.180 & 3.598 \\ \hline
			CFS & 5.283 & 0.505 & 5.365 & 0.125 & 10.179 & 11.123 & 4.654 \\ \hline
			Trace-Ratio & 1.351 & 0.080 & 2.792 & 0.038 & 16.530 & 16.382 & 5.310 \\ \hline
			HSIC-Lasso & 28.631 & 25.108 & 189.633 & 0.383 & 230.136 & 247.173 & 103.009 \\ \hline
			LassoNet & 115.466 & 67.480 & 151.358 & 19.937 & 565.827 & 490.660 & 201.533 \\ \hline
			CMIM & 1536.333 & 147.626 & 2345.528 & 5.599 & 5684.953 & 6688.953 & 2344.142 \\ \hline
			mRMR & 1583.584 & 148.563 & 2399.907 & 5.727 & 5767.651 & 6705.513 & 2372.992 \\ \hline
			JMI & 1553.416 & 147.447 & 2435.854 & 5.585 & 5989.872 & 6666.467 & 2399.806 \\ \hline
			UDFS & 235.908 & 5.108 & 579.341 & 1.860 & 9558.169 & 9838.824 & 2888.458 \\ \hline
			DISR & 4269.103 & 399.536 & 6489.994 & 15.520 & 15525.915 & 17810.256 & 6358.618 \\ \hline
		\end{tabular}
	\end{adjustbox}
	\caption{Time consuming of feature selection.}
	\label{fs_time_consuming.}
\end{table}

	Our method has a huge advantage in terms of running speed, thus it can be used in machine learning workflows with very low cost, which is different from most of the existing methods.
	As shown in Figure \ref{log_unit_gain_vs_acc}, considering the classic machine learning pipeline that feature selection followed by model training, our method has the lowest time cost per unit accuracy on all benchmark datasets.
	Compared to the almost straight logarithmic cost curves of other methods, the curve of our method has a more pronounced curvature, which means that the time required by our method is almost negligible relative to the overall process.
	Given the time saved in the feature selection procedure, one could try more features in training and inference to achieve higher performance.
	In a nutshell, ContrastFS achieves the best trade-off between effectiveness and efficiency in this comparison.

\begin{figure}[!h]
	\centering
	\begin{subfigure}[b]{0.32\textwidth}
		\includegraphics[width=\textwidth]{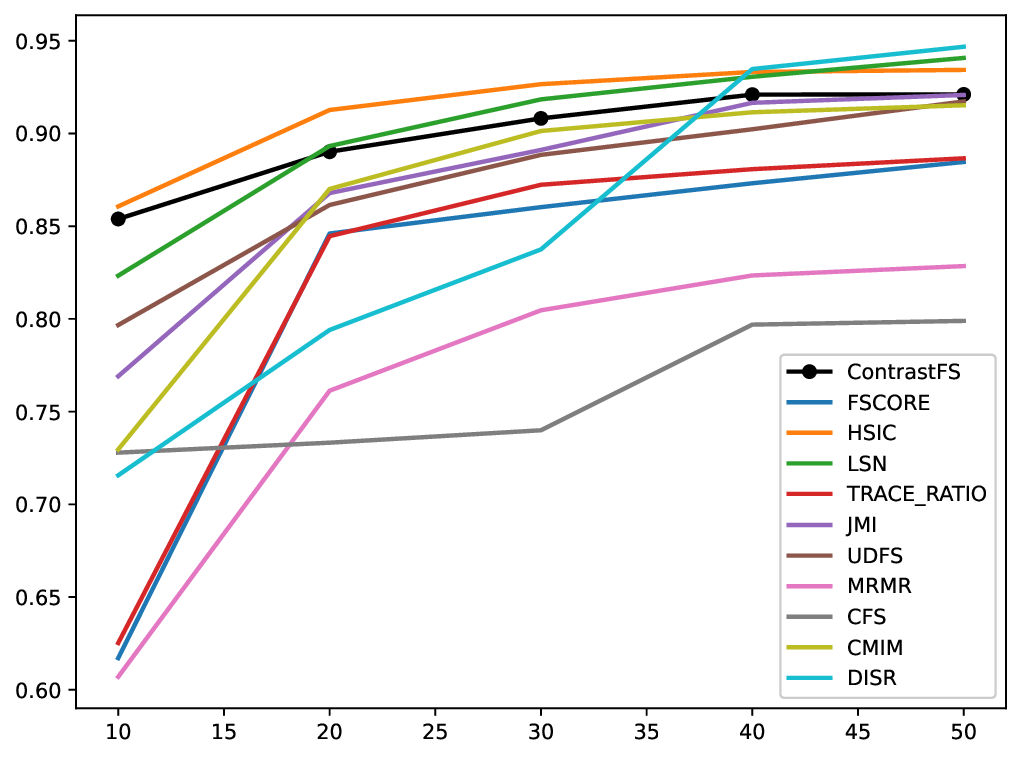} 
		\caption{Activity}
	\label{Activity_acc_xgb}    
	\end{subfigure}
	\begin{subfigure}[b]{0.32\textwidth}
		\includegraphics[width=\textwidth]{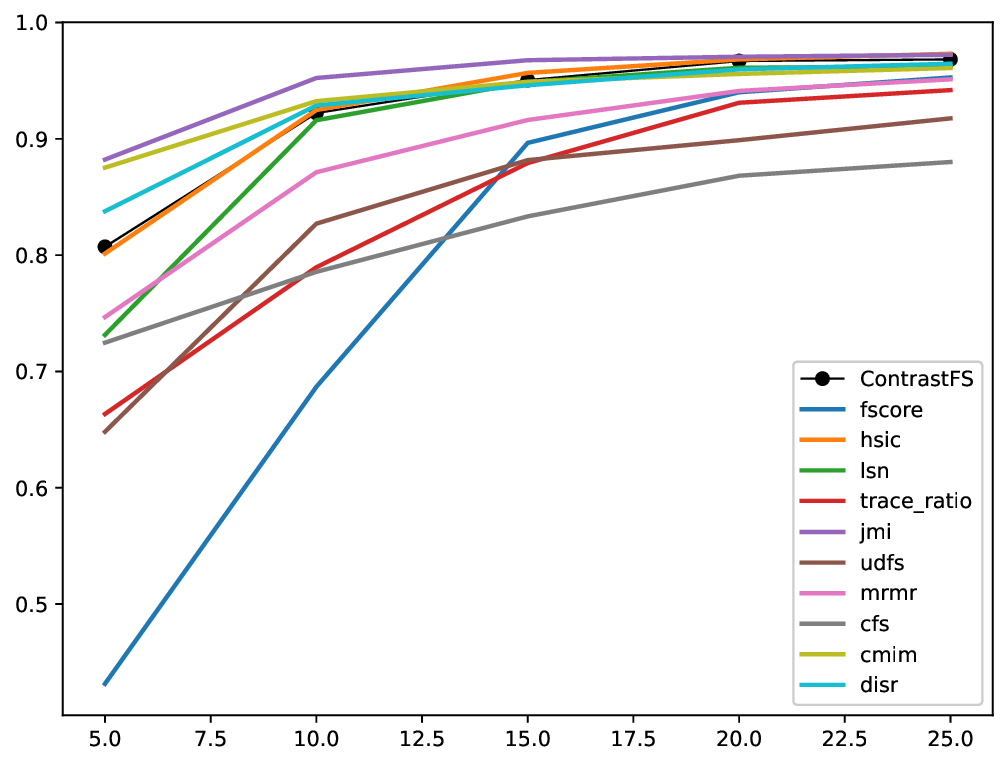}
		\caption{COIL-20}
		\label{COIL-20_acc_xgb}    
	\end{subfigure}
	\begin{subfigure}[b]{0.32\textwidth}
		\includegraphics[width=\textwidth]{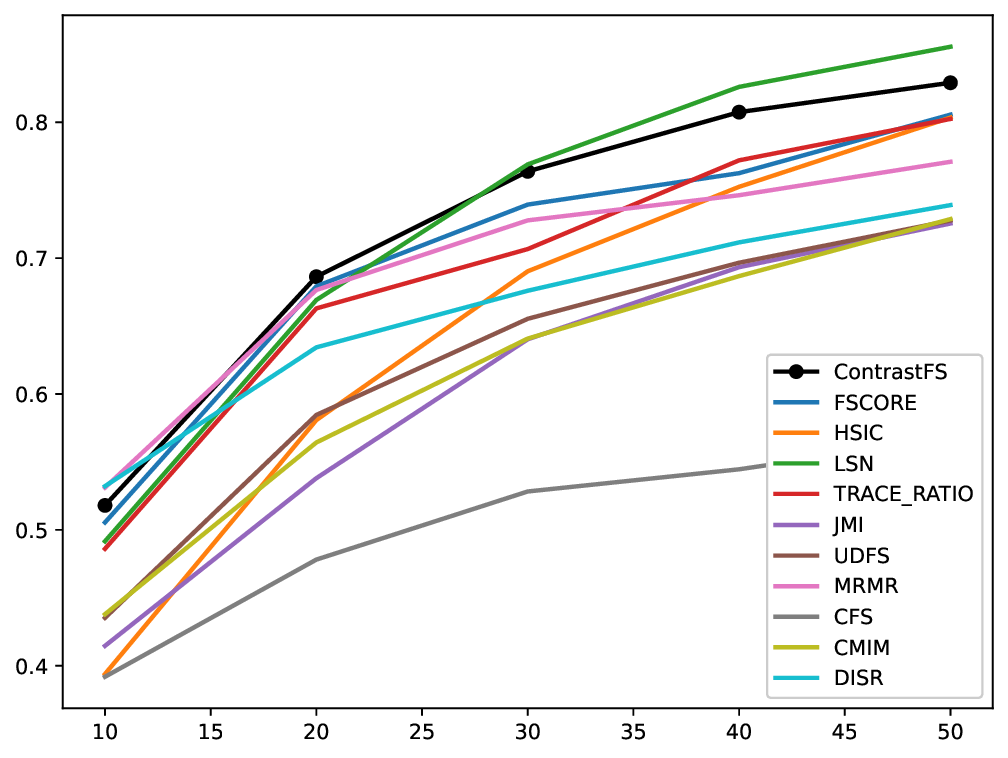}
		\caption{ISOLET}
		\label{ISOLET_acc_xgb}    
	\end{subfigure}
	~ 
	\begin{subfigure}[b]{0.32\textwidth}
		\includegraphics[width=\textwidth]{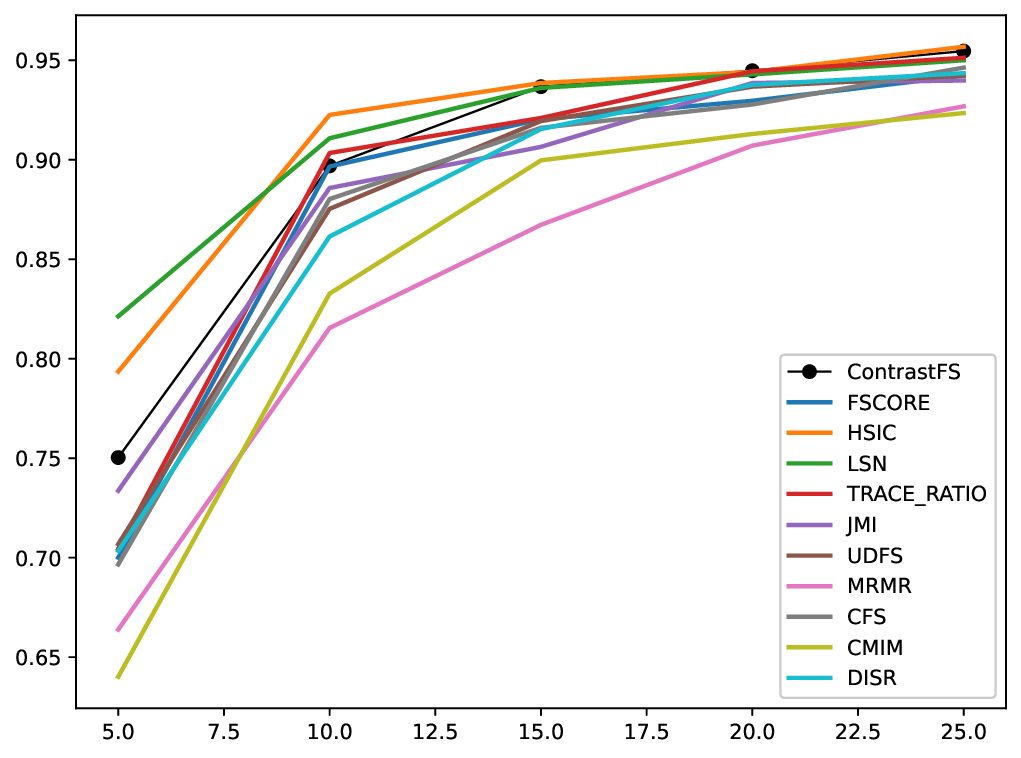}
		\caption{MICE}
		\label{MICE_acc_xgb}    
	\end{subfigure}
	\begin{subfigure}[b]{0.32\textwidth}
		\includegraphics[width=\textwidth]{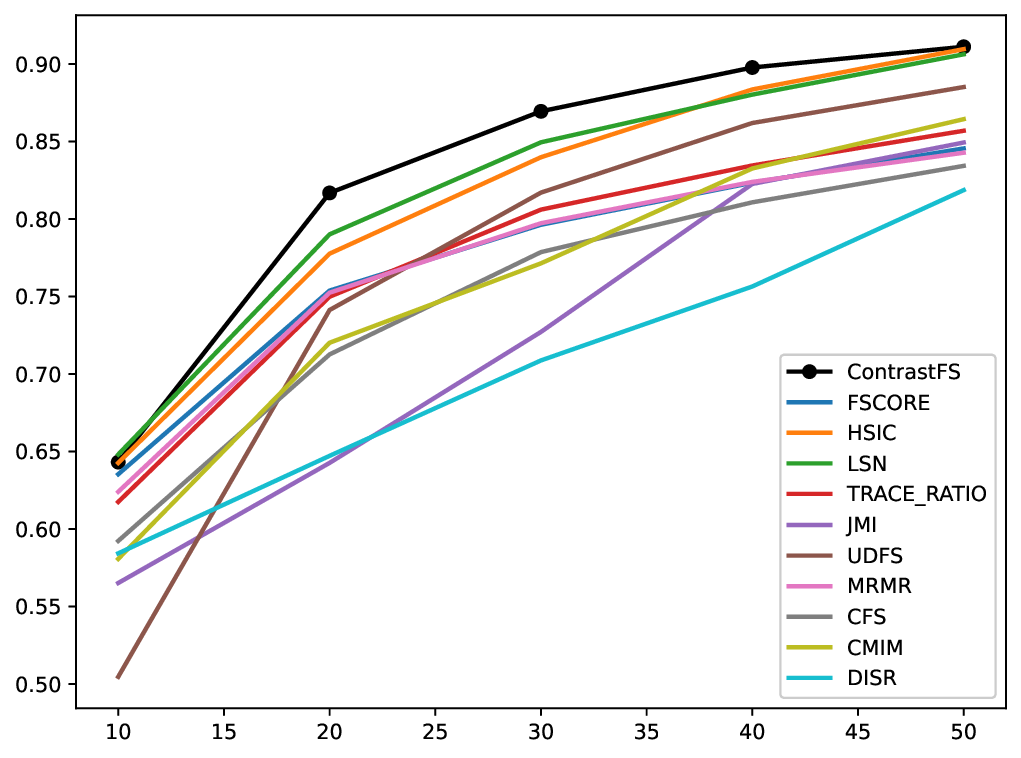}
		\caption{MNIST}
		\label{MNIST_acc_xgb}    
	\end{subfigure}
	\begin{subfigure}[b]{0.32\textwidth}
		\includegraphics[width=\textwidth]{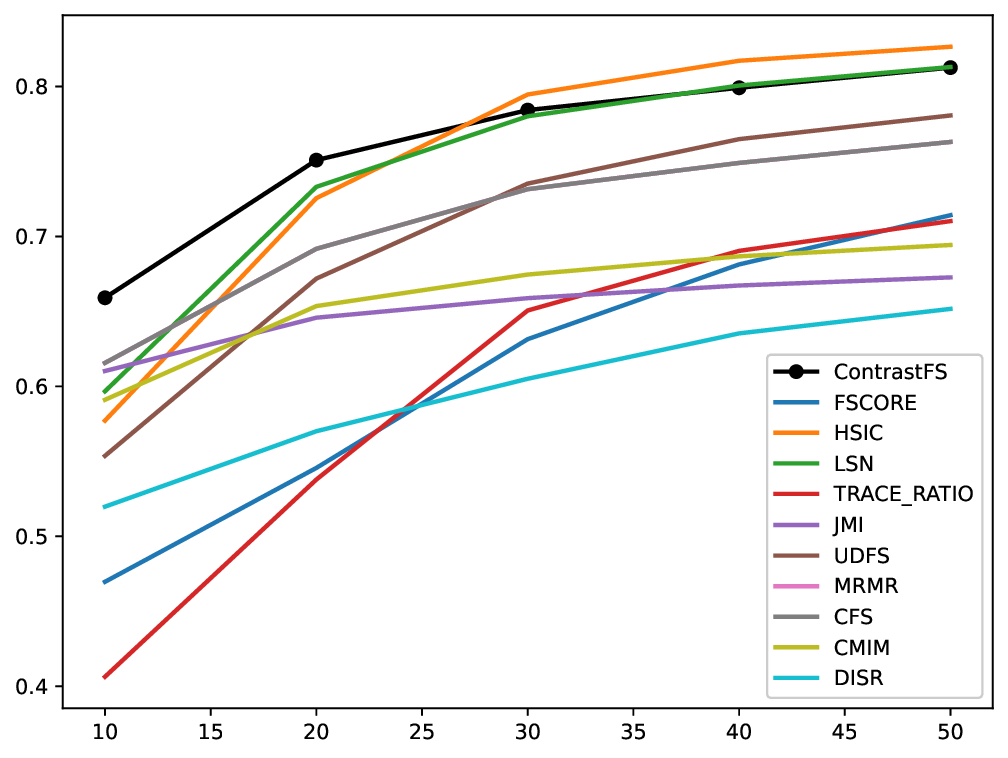}
		\caption{Fashion-MNIST}
		\label{MNIST-Fashion_acc_xgb}    
	\end{subfigure}
	\caption{Classification accuracy vs. number of selected features. This figure shows the average accuracy achieved by XGBoost classifieds fed in features selected by various methods in multiple runs on benchmark datasets.}
	\label{acc_vs_fs_num.}
\end{figure}
	
	\subsection{Classification accuracy}
	
	The accuracy of classification using a set of features is a natural way to assess the value of the features in supervised learning, and it is also commonly used to compare the performance of feature selection methods.
	We use this metric to evaluate different methods by varying the number of features they select.
	For the Mice and COIL-20 datasets, which are small in size, we test the accuracy of feature sets with five different sizes, ranging from 5 to 25. For the other datasets, we test the feature sets with sizes, ranging from 10 to 50.
	We choose XGBoost \cite{chen2016xgboost} as the downstream classifier, for its advantages in performance, interpretation, and ease to control.
	We implement the classifier using Scikit-learn \cite{pedregosa2011scikitlearn} and the xgboost python package. For fair fairness, we set the same hyperparameter for all feature sets, set the number of trees to 50, and set the random state of the classifiers to 0.

	Our method exhibits superior performance on various datasets, ranging from pattern recognition to tabular data, by achieving higher accuracy than other filter methods in most scenarios, and generally comparable to LassoNet and HSIC-Lasso. 
	On the COIL-20, ISOLET and Fashion-MNIST datasets, ContrastFS outperforms HSIC-Lasso and LassoNet at some stages.
	On the MNIST dataset, our method consistently outperforms other methods throughout the entire process.
	Even on datasets with small data size, such as Mice and COIL-20, our method still shows significant advantages over other filter methods.
	Notably, our method performs excellently on datasets from different domains and types, which indicates that our method has a broad applicability.

	\subsection{Reduce the Redundancy}
	Unlike the common way to explore redundancy in features by analyzing the original data, we introduce a method based on discrepancies features shown in different surrogate representations. 
	As introduced in Section \ref{reduce_redundancy}, we can measure the correlation between features and eliminate the highly correlated ones from the selected feature subset based on discrepancies shown by surrogate representation in different classes, then reduce the redundancy by adopting a simple strategy: filter out the highly correlated features.
	We implement experiments on several datasets to validate its effectiveness.
	
\begin{figure}[!t]
		\centering
		\begin{subfigure}[b]{0.32\textwidth}
			\includegraphics[width=\textwidth]{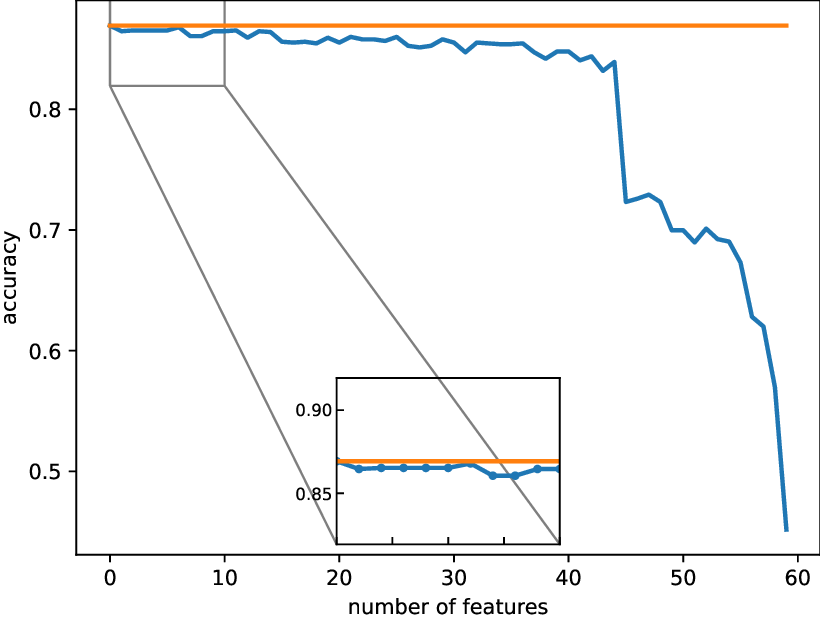} 
			\caption{Activity}
		\end{subfigure}
		\begin{subfigure}[b]{0.32\textwidth}
			\includegraphics[width=\textwidth]{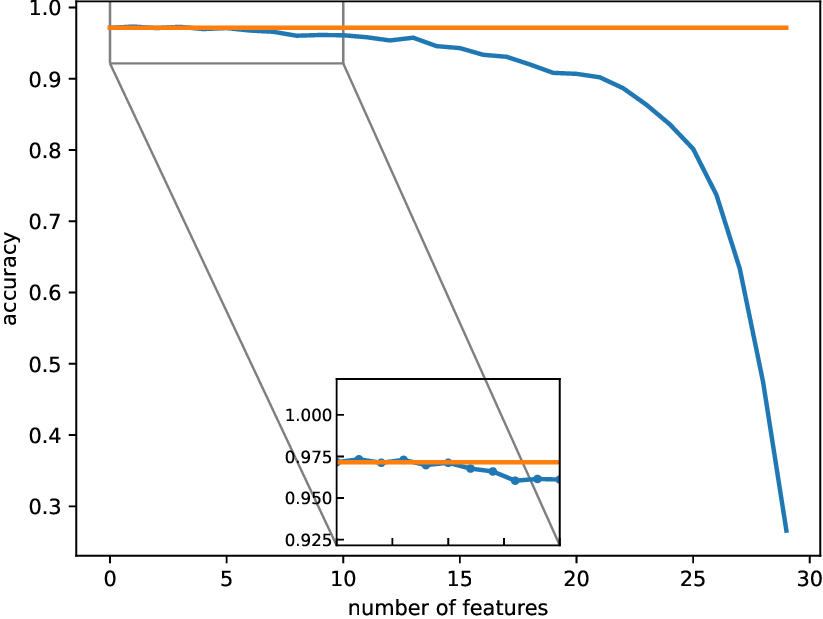}
			\caption{COIL-20}
		\end{subfigure}
		\begin{subfigure}[b]{0.32\textwidth}
			\includegraphics[width=\textwidth]{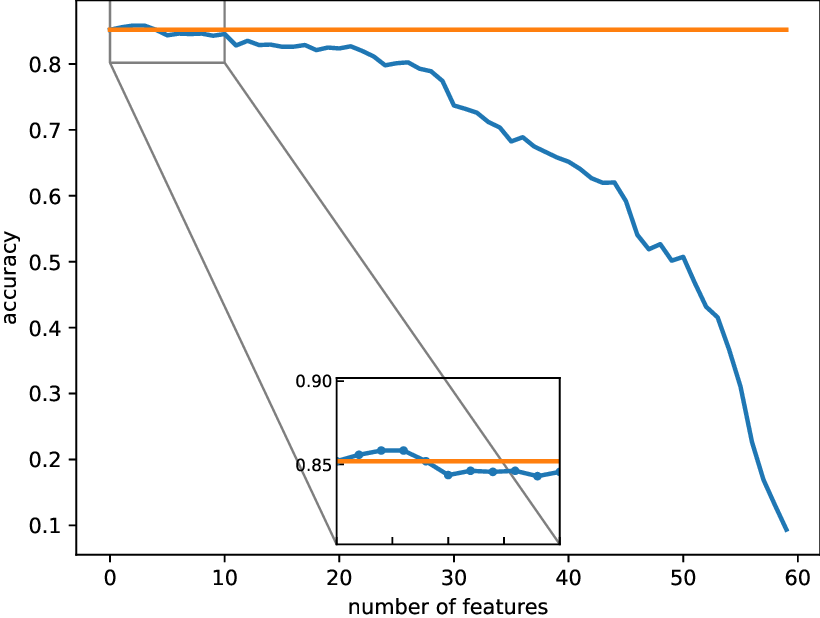}
			\caption{ISOLET}
		\end{subfigure}
		\begin{subfigure}[b]{0.32\textwidth}
			\includegraphics[width=\textwidth]{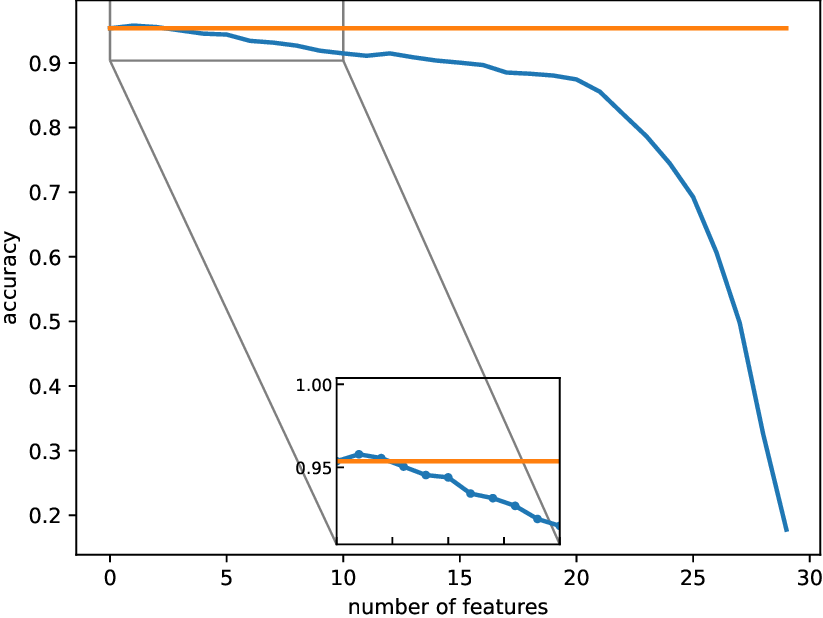}
			\caption{Mice}
		\end{subfigure}
		\begin{subfigure}[b]{0.32\textwidth}
			\includegraphics[width=\textwidth]{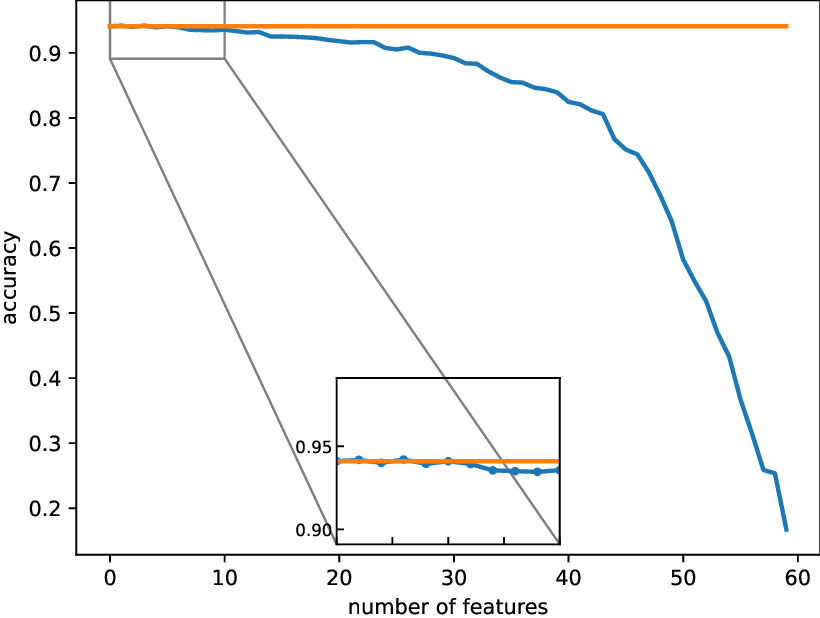}
			\caption{MNIST}
		\end{subfigure}
		\begin{subfigure}[b]{0.32\textwidth}
			\includegraphics[width=\textwidth]{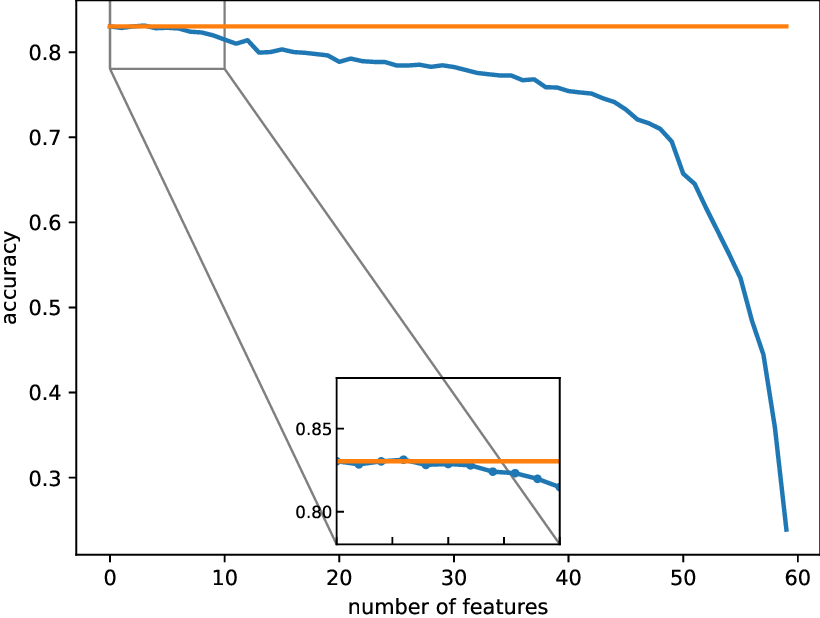}
			\caption{Fashion-MNIST}
		\end{subfigure}
		
		\caption{Accuracy of the feature subset after redundancy reduction. This figure shows the accuracy of XGBoost with a feature subset that starting from 60 features (30 features for COIL-20 or MICE), selected by our method and reducing the redundancy according to the correlation between the surrogate representations.}
		\label{acc_decor_drop}
	\end{figure}

	We use the ContrastFS method to select a feature set T from the benchmark datasets. The size of T is 60 for the Activity, ISOLET, MNIST, and Fashion-MNIST datasets, and 30 for the COIL-20 and Mice datasets.
	We then compute the correlation between these features and use the row-wise average of the correlation matrix as the redundancy measure for each feature, remove the features from T in descending order of their redundancy and use the remaining feature set for classification. The accuracy path is shown in Figure \ref{acc_decor_drop}. 
	For the Activity dataset, the accuracy does not decrease significantly when we remove the first 10 features. However, it declines sharply when the feature set has less than 20 features. For the other datasets, the accuracy improves when we eliminate some redundant features, especially for the ISOLET dataset.
	These results further confirm the feasibility of using surrogate representations to study the utility of features and their relationships.
	
	\subsection{Stablility}
	
\begin{figure}[!t]
	\centering
	\begin{subfigure}[b]{0.32\textwidth}
		\includegraphics[width=\textwidth]{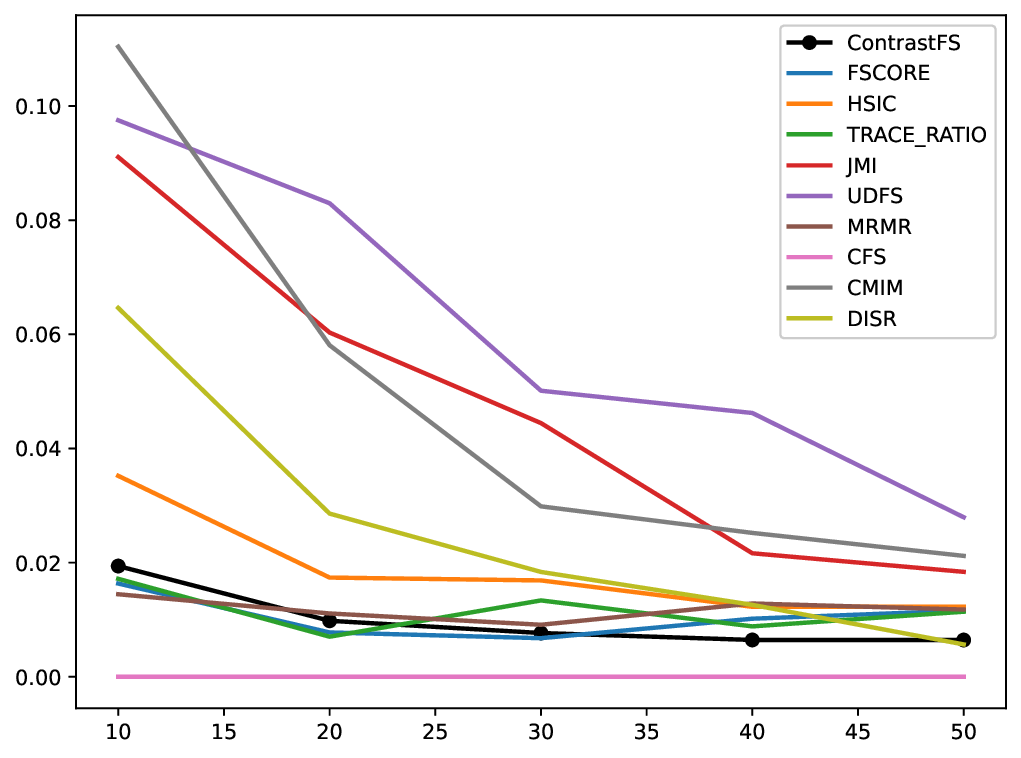} 
		\caption{Activity}
	\end{subfigure}
	\begin{subfigure}[b]{0.32\textwidth}
		\includegraphics[width=\textwidth]{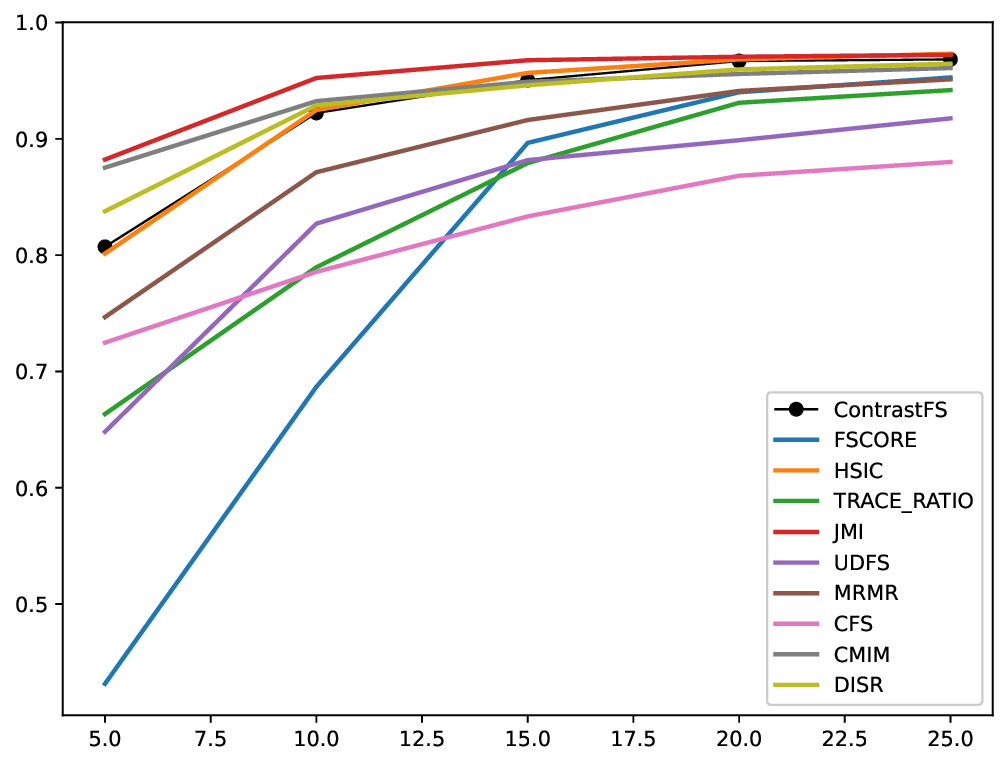}
		\caption{COIL-20}
	\end{subfigure}
	\begin{subfigure}[b]{0.32\textwidth}
		\includegraphics[width=\textwidth]{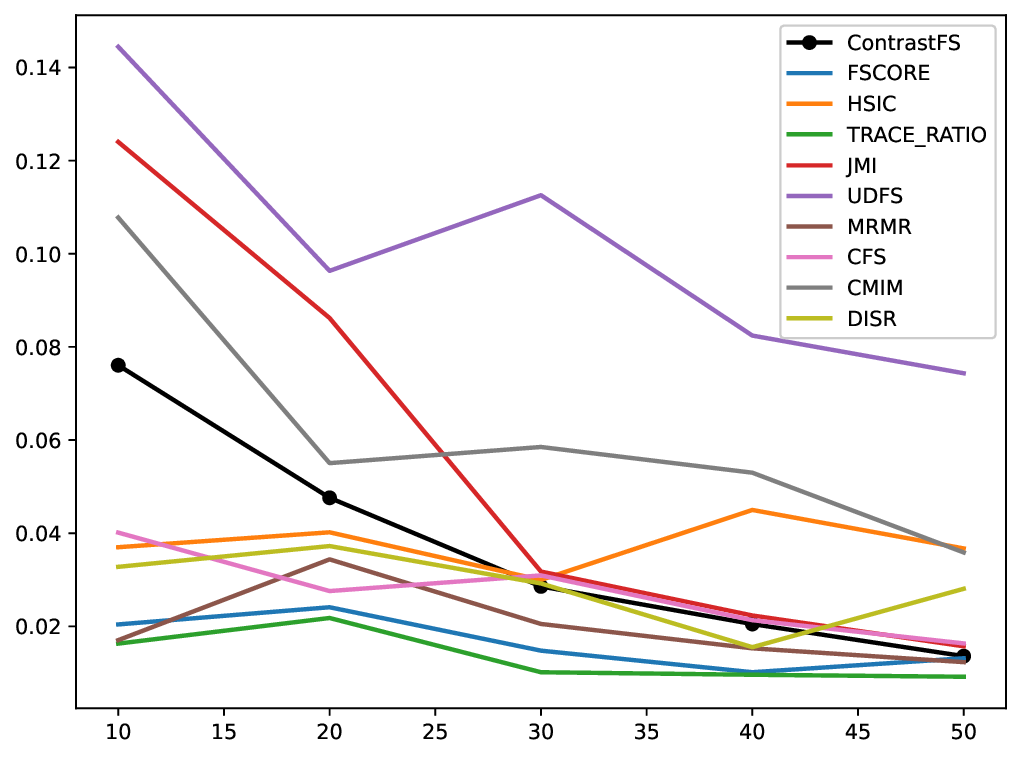}
		\caption{ISOLET}
	\end{subfigure}
	\begin{subfigure}[b]{0.32\textwidth}
		\includegraphics[width=\textwidth]{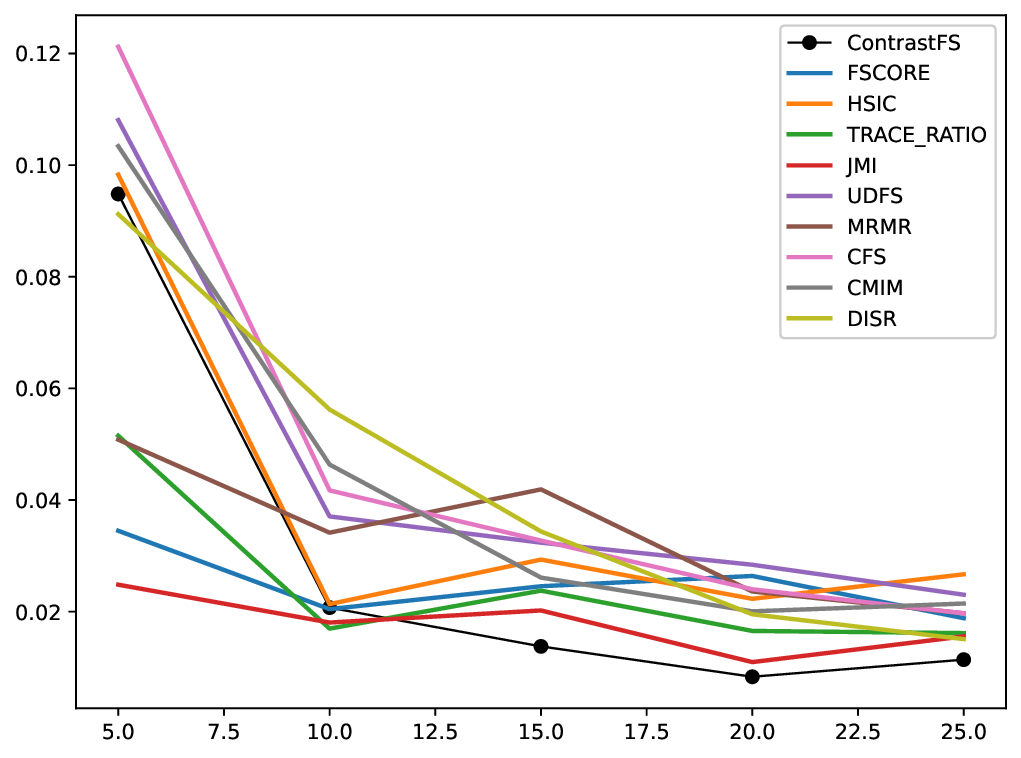}
		\caption{Mice}
	\end{subfigure}
	\begin{subfigure}[b]{0.32\textwidth}
		\includegraphics[width=\textwidth]{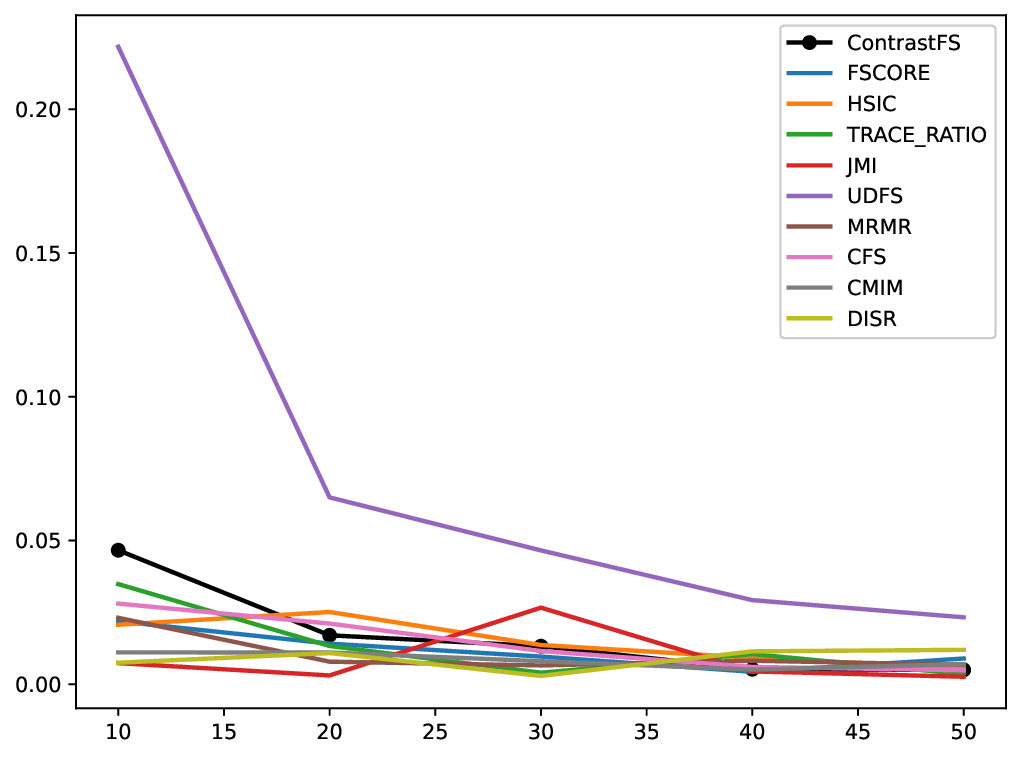}
		\caption{MNIST}
	\end{subfigure}
	\begin{subfigure}[b]{0.32\textwidth}
		\includegraphics[width=\textwidth]{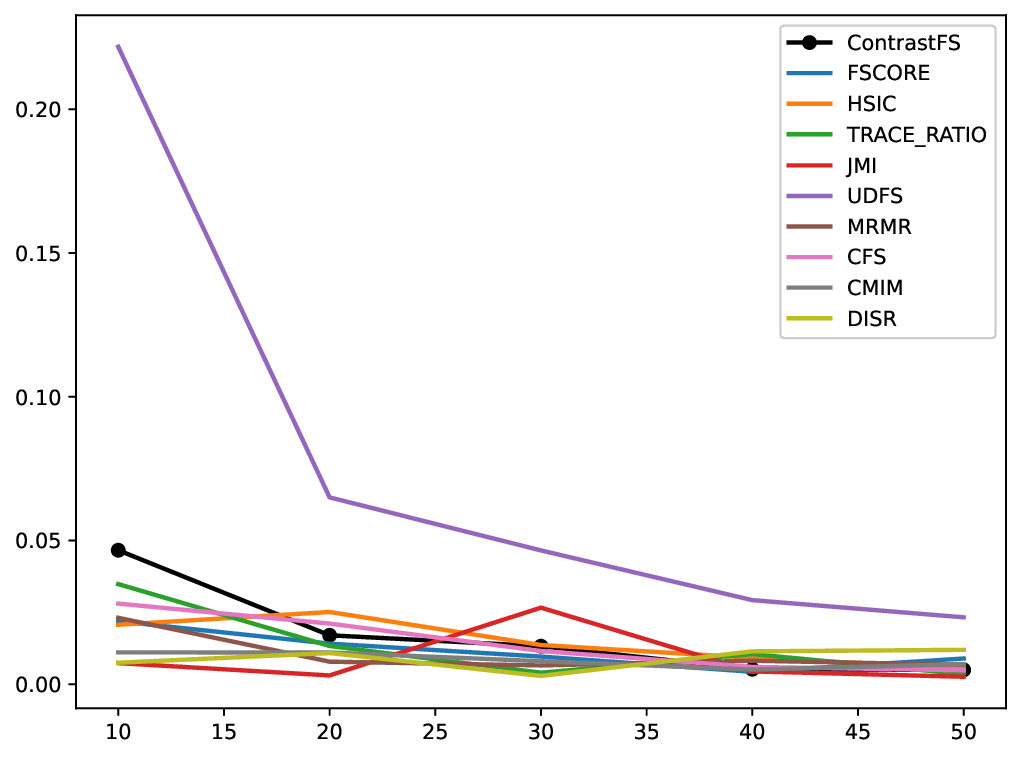}
		\caption{Fashion-MNIST}
	\end{subfigure}
	\caption{Relative standard deviation of accuracy.}
	\label{Relative_standard_deviation_of_accuracy}
\end{figure}

	The stability of feature selection methods can be evaluated from two perspectives: one is their robustness to sample perturbations, and the other is their resilience to noise in the data. We perform experiments on both aspects to demonstrate the stability of our method.
	We run feature selection methods in comparison \footnote{except LassoNet, for the uncertainty of its integrated backbone model.} 10 times on benchmark datasets to study the stability of our method when faced with instance-wise perturbation, in each run we randomly split datasets with predefined random state to ensure that each method faces the same data, and the random state numbers are chosen randomly.
	We consider the relative standard deviation (RSD) of accuracy as a tool to quantify the stability of feature selection methods, it's the ratio of standard error of accuracy and accuracy.
	$$
	RSD = \frac{Std(Accuracy) }{Accuracy}
	$$
	As shown in Figure \ref{Relative_standard_deviation_of_accuracy}, our method shows excellent stability on datasets other than ISOLET, and achieves the lowest RSD of accuracy on the Activity and Mice datasets at some points, although not throughout the whole process. It also performs close to the best on the MNIST, Fashion-MNIST and COIL-20 datasets.
	Even on the ISOLET dataset, our method is not far behind the best one at that time.
	The results show that our method has good stability for instance-wise perturbations.
	
\begin{figure}[]
	\centering
	
	\begin{subfigure}[b]{0.48\textwidth}
		\includegraphics[width=\textwidth]{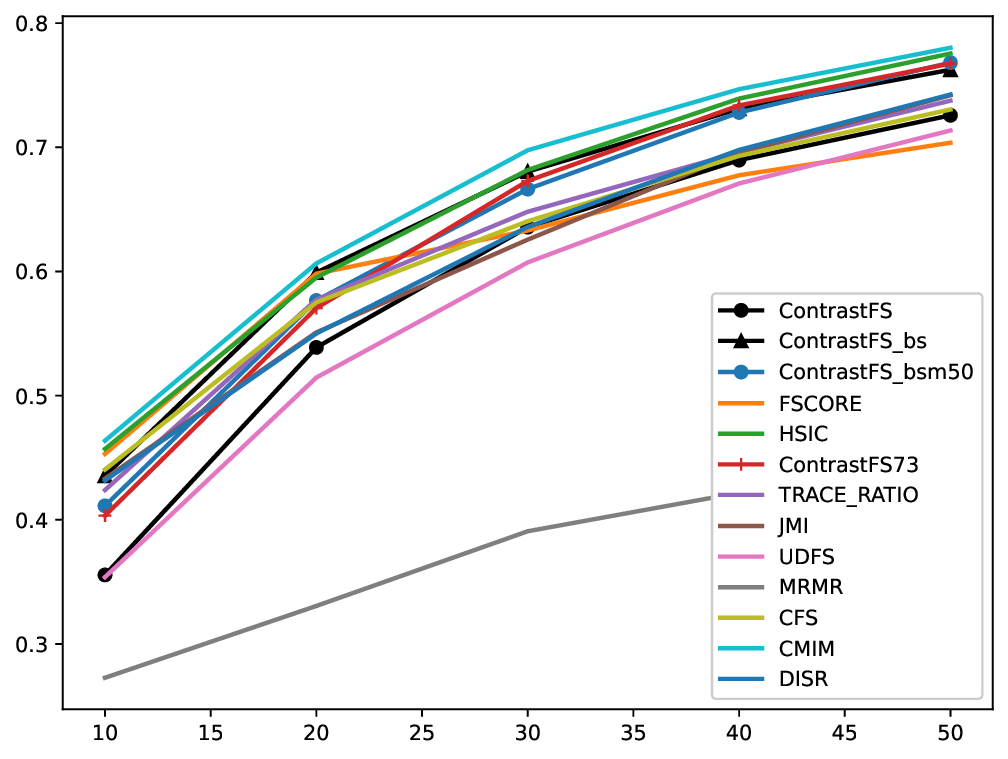}
		\caption{Accuracy}
	\end{subfigure}
	\begin{subfigure}[b]{0.48\textwidth}
		\includegraphics[width=\textwidth]{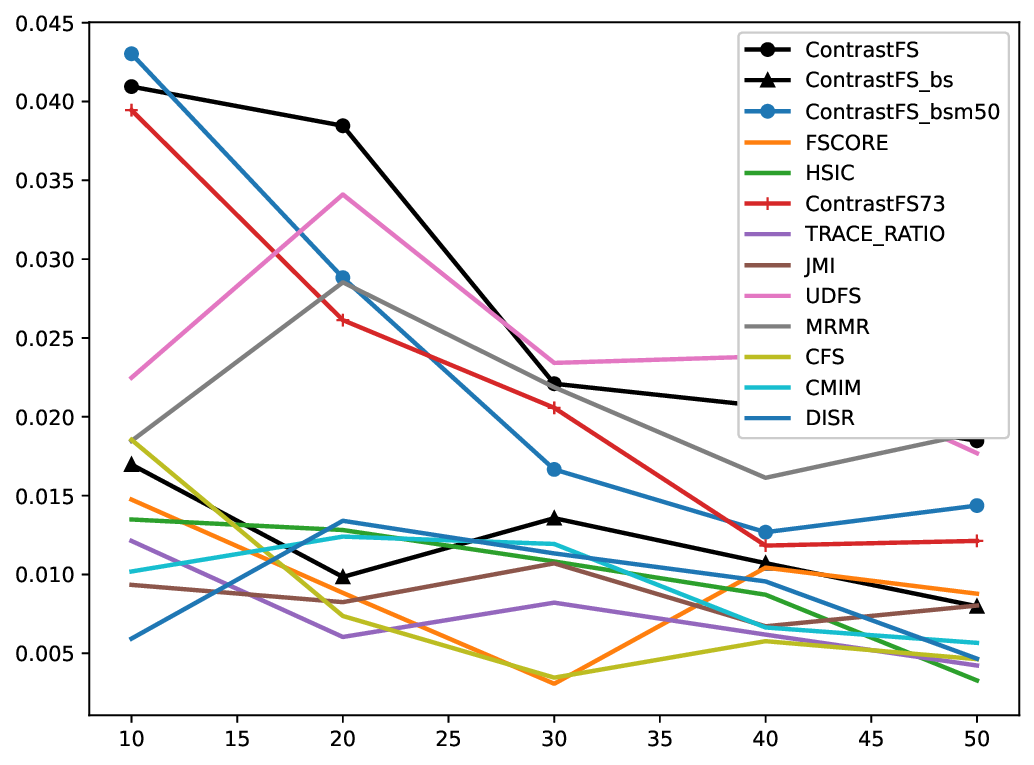}
		\caption{Stndard deviation of accuracy}
	\end{subfigure}
	\caption{Accuracy improvement using the bootstrap method. (a) The mean of the accuracy achieved by our methods. (b) The standard deviation of the accuracy achieved by our methods.}
	\label{bootstrap_on_nmnit}
\end{figure}

	The noise in the data requires more samples to obtain a reliable estimate of a given accuracy level, which poses some challenges for our method. Figure \ref{bootstrap_on_nmnit} illustrates this point: when only 20\% of the nMNIST-AWGN dataset is used for feature evaluation, ContrastFS performs poorly, while using the same 70-30 split as other tabular data, it performs nearly optimal. Since analysis in practice is often based on limited noisy data, we suggest using bootstrap to improve the performance of our method. Our method has a significant speed advantage over other methods, and the overhead of using bootstrap is low, while the performance gain is substantial. We also note that bootstrapping the sample moments and applying them to our method is a viable option, but it is slightly inferior to bootstrapping the importance score derived by ContrastFS in terms of both accuracy and stability.

	\section{Conclusion}
	
	In this paper, we propose a new feature selection method, ContrastFS, which selects important features based on the discrepancies features to show between classes. 
	Unlike most other feature selection methods, our method analyzes features based on surrogate representations rather than directly on the original dataset.
	At its core, ContrastFS constructs surrogate representations of classes based on the low-order moments estimated from datasets, and then merits features according to the discrepancies between classes. 
	The estimation of low-order moments requires little computational effort, and the surrogate representations get the further analysis of feature behavior rid of original high-dimensional datasets, resulting in the high efficiency of ContrastFS.
	
	The advantages of ContrastFS are that it does not rely on complicated assumptions about the datasets and it works independently of learning models, which makes it general and easy to use. 
	ContrastFS could work with most common learning models and could be a costless preprocessing procedure in the machine learning pipeline, and could also be used in various domains.
	The most valuable characteristic of ContrastFS is the high efficiency it possesses, we can implement it with negligible cost in various devices since the components of it are all basic operations and could be vectorized and performed in parallel.
	
	ContrastFS evaluates features individually; like most of the other filter feature selection methods we've compared in this paper, it's also possible to explore the relationship and reduce redundancy based on surrogate representations, which could lead to higher performance.

	The Python code and documentation for ContrastFS are available at \textbf{github url}.
	
	\subsection*{Acknowledgments}
	
	\newpage
	
	\bibliography{feature_selection_bibt.bib}

\begin{thebibliography}{41}
\providecommand{\natexlab}[1]{#1}
\providecommand{\url}[1]{\texttt{#1}}
\expandafter\ifx\csname urlstyle\endcsname\relax
  \providecommand{\doi}[1]{doi: #1}\else
  \providecommand{\doi}{doi: \begingroup \urlstyle{rm}\Url}\fi

\bibitem[Bal{\i}n et~al.(2019)Bal{\i}n, Abid, and Zou]{balin2019concrete}
Bal{\i}n, M.~F., Abid, A., and Zou, J.
\newblock Concrete autoencoders: {{Differentiable}} feature selection and
  reconstruction.
\newblock In \emph{International Conference on Machine Learning}, pp.\
  444--453. {PMLR}, 2019.

\bibitem[Bellman(1957)]{bellman1957dynamic}
Bellman, R.
\newblock \emph{Dynamic Programming}.
\newblock {Princeton University Press}, {Princeton}, 1957.

\bibitem[{Bol{\'o}n-Canedo} \& Remeseiro(2020){Bol{\'o}n-Canedo} and
  Remeseiro]{bolon-canedo2020feature}
{Bol{\'o}n-Canedo}, V. and Remeseiro, B.
\newblock Feature selection in image analysis: A survey.
\newblock \emph{Artificial Intelligence Review}, 53\penalty0 (4):\penalty0
  2905--2931, April 2020.

\bibitem[Brown et~al.(2012)Brown, Pocock, Zhao, and
  Luj{\'a}n]{brown2012conditional}
Brown, G., Pocock, A., Zhao, M.-J., and Luj{\'a}n, M.
\newblock Conditional {{Likelihood Maximisation}}: {{A Unifying Framework}} for
  {{Information Theoretic Feature Selection}}.
\newblock \emph{Journal of Machine Learning Research}, 13\penalty0
  (2):\penalty0 27--66, 2012.

\bibitem[Cai et~al.(2018)Cai, Luo, Wang, and Yang]{cai2018feature}
Cai, J., Luo, J., Wang, S., and Yang, S.
\newblock Feature selection in machine learning: {{A}} new perspective.
\newblock \emph{Neurocomputing}, 300:\penalty0 70--79, July 2018.

\bibitem[Chandrashekar \& Sahin(2014)Chandrashekar and
  Sahin]{chandrashekar2014survey}
Chandrashekar, G. and Sahin, F.
\newblock A survey on feature selection methods.
\newblock \emph{Computers \& Electrical Engineering}, 40\penalty0 (1):\penalty0
  16--28, January 2014.

\bibitem[Chen \& Guestrin(2016)Chen and Guestrin]{chen2016xgboost}
Chen, T. and Guestrin, C.
\newblock {{XGBoost}}: {{A Scalable Tree Boosting System}}.
\newblock In \emph{Proceedings of the 22nd {{ACM SIGKDD International
  Conference}} on {{Knowledge Discovery}} and {{Data Mining}}}, {{KDD}} '16,
  pp.\  785--794, {New York, NY, USA}, August 2016. {Association for Computing
  Machinery}.
\newblock ISBN 978-1-4503-4232-2.

\bibitem[{Climente-Gonz{\'a}lez} et~al.(2019){Climente-Gonz{\'a}lez}, Azencott,
  Kaski, and Yamada]{climente-gonzalez2019block}
{Climente-Gonz{\'a}lez}, H., Azencott, C.-A., Kaski, S., and Yamada, M.
\newblock Block {{HSIC Lasso}}: Model-free biomarker detection for ultra-high
  dimensional data.
\newblock \emph{Bioinformatics}, 35\penalty0 (14):\penalty0 i427--i435, July
  2019.

\bibitem[Dess{\`i} \& Pes(2015)Dess{\`i} and Pes]{dessi2015similarity}
Dess{\`i}, N. and Pes, B.
\newblock Similarity of feature selection methods: {{An}} empirical study
  across data intensive classification tasks.
\newblock \emph{Expert Systems with Applications}, 42\penalty0 (10):\penalty0
  4632--4642, June 2015.

\bibitem[Duda et~al.(2000)Duda, Hart, and Stork]{duda2000pattern}
Duda, R.~O., Hart, P.~E., and Stork, D.~G.
\newblock \emph{Pattern Classification}.
\newblock {Wiley}, {New York}, 2nd ed edition, 2000.
\newblock ISBN 978-0-471-05669-0.

\bibitem[Fan \& Li(2006)Fan and Li]{fan2006statistical}
Fan, J. and Li, R.
\newblock {Statistical challenges with high dimensionality: Feature selection
  in knowledge discovery}.
\newblock In \emph{{25th International Congress of Mathematicians, ICM 2006}},
  2006.

\bibitem[Fleuret(2004)]{fleuret2004fast}
Fleuret, F.
\newblock Fast {{Binary Feature Selection}} with {{Conditional Mutual
  Information}}.
\newblock \emph{Journal of Machine Learning Research}, 5:\penalty0 1531--1555,
  December 2004.

\bibitem[Goodfellow et~al.(2016)Goodfellow, Bengio, and
  Courville]{goodfellow2016deep}
Goodfellow, I., Bengio, Y., and Courville, A.
\newblock \emph{Deep Learning}.
\newblock Adaptive Computation and Machine Learning. {The MIT Press},
  {Cambridge, Massachusetts}, 2016.
\newblock ISBN 978-0-262-03561-3.

\bibitem[Guyon \& Elisseeff(2003)Guyon and Elisseeff]{guyon2003introduction}
Guyon, I. and Elisseeff, A.
\newblock An introduction to variable and feature selection.
\newblock \emph{Journal of machine learning research}, 3\penalty0
  (Mar):\penalty0 1157--1182, 2003.

\bibitem[Hall \& Smith(1999)Hall and Smith]{hall1999feature}
Hall, M.~A. and Smith, L.~A.
\newblock Feature selection for machine learning: Comparing a correlation-based
  filter approach to the wrapper.
\newblock In \emph{{{FLAIRS}} Conference}, volume 1999, pp.\  235--239, 1999.

\bibitem[Hancer et~al.(2020)Hancer, Xue, and Zhang]{hancer2020survey}
Hancer, E., Xue, B., and Zhang, M.
\newblock A survey on feature selection approaches for clustering.
\newblock \emph{Artificial Intelligence Review}, 53\penalty0 (6):\penalty0
  4519--4545, August 2020.

\bibitem[{Hanchuan Peng} et~al.(2005){Hanchuan Peng}, {Fuhui Long}, and
  Ding]{hanchuanpeng2005feature}
{Hanchuan Peng}, {Fuhui Long}, and Ding, C.
\newblock Feature selection based on mutual information criteria of
  max-dependency, max-relevance, and min-redundancy.
\newblock \emph{IEEE Transactions on Pattern Analysis and Machine
  Intelligence}, 27\penalty0 (8):\penalty0 1226--1238, August 2005.

\bibitem[Johnstone \& Titterington(2009)Johnstone and
  Titterington]{johnstone2009statistical}
Johnstone, I.~M. and Titterington, D.~M.
\newblock Statistical challenges of high-dimensional data.
\newblock \emph{Philosophical Transactions of the Royal Society A:
  Mathematical, Physical and Engineering Sciences}, 367\penalty0
  (1906):\penalty0 4237--4253, November 2009.

\bibitem[Kohavi \& John(1997)Kohavi and John]{kohavi1997wrappers}
Kohavi, R. and John, G.~H.
\newblock Wrappers for feature subset selection.
\newblock \emph{Artificial Intelligence}, 97\penalty0 (1-2):\penalty0 273--324,
  December 1997.

\bibitem[Konda et~al.(2013)Konda, Kumar, R{\'e}, and
  Sashikanth]{konda2013feature}
Konda, P., Kumar, A., R{\'e}, C., and Sashikanth, V.
\newblock Feature selection in enterprise analytics: A demonstration using an
  {{R-based}} data analytics system.
\newblock \emph{Proceedings of the VLDB Endowment}, 6\penalty0 (12):\penalty0
  1306--1309, August 2013.

\bibitem[Lemhadri et~al.(2021)Lemhadri, Ruan, Abraham, and
  Tibshirani]{lemhadri2021lassonet}
Lemhadri, I., Ruan, F., Abraham, L., and Tibshirani, R.
\newblock {{LassoNet}}: {{A Neural Network}} with {{Feature Sparsity}}.
\newblock \emph{Journal of Machine Learning Research}, 22\penalty0 (127), June
  2021.

\bibitem[Li \& Liu(2017)Li and Liu]{li2017challenges}
Li, J. and Liu, H.
\newblock Challenges of {{Feature Selection}} for {{Big Data Analytics}}.
\newblock \emph{IEEE Intelligent Systems and Their Applications}, 32\penalty0
  (2):\penalty0 9--15, March 2017.

\bibitem[Li et~al.(2018)Li, Cheng, Wang, Morstatter, Trevino, Tang, and
  Liu]{li2018feature}
Li, J., Cheng, K., Wang, S., et~al.
\newblock Feature {{Selection}}: {{A Data Perspective}}.
\newblock \emph{ACM Computing Surveys}, 50\penalty0 (6):\penalty0 1--45,
  November 2018.

\bibitem[Li et~al.(2017)Li, Li, and Liu]{li2017recent}
Li, Y., Li, T., and Liu, H.
\newblock Recent advances in feature selection and its applications.
\newblock \emph{Knowledge and Information Systems}, 53\penalty0 (3):\penalty0
  551--577, December 2017.

\bibitem[Meyer et~al.(2008)Meyer, Schretter, and
  Bontempi]{meyer2008informationtheoretic}
Meyer, P.~E., Schretter, C., and Bontempi, G.
\newblock Information-{{Theoretic Feature Selection}} in {{Microarray Data
  Using Variable Complementarity}}.
\newblock \emph{IEEE Journal of Selected Topics in Signal Processing},
  2\penalty0 (3):\penalty0 261--274, June 2008.

\bibitem[Nie et~al.(2008)Nie, Xiang, Jia, Zhang, and Yan]{nie2008trace}
Nie, F., Xiang, S., Jia, Y., Zhang, C., and Yan, S.
\newblock Trace ratio criterion for feature selection.
\newblock In \emph{{{AAAI}}}, volume~2, pp.\  671--676, 2008.

\bibitem[Nie et~al.(2010)Nie, Huang, Cai, and Ding]{nie2010efficient}
Nie, F., Huang, H., Cai, X., and Ding, C.
\newblock Efficient and {{Robust Feature Selection}} via {{Joint}}
  \textbackslash mathscrl2,1-{{Norms Minimization}}.
\newblock In \emph{Advances in {{Neural Information Processing Systems}}},
  volume~23. {Curran Associates, Inc.}, 2010.

\bibitem[Pedregosa et~al.(2011)Pedregosa, Varoquaux, Gramfort, Michel, Thirion,
  Grisel, Blondel, Prettenhofer, Weiss, Dubourg, Vanderplas, Passos,
  Cournapeau, Brucher, Perrot, and Duchesnay]{pedregosa2011scikitlearn}
Pedregosa, F., Varoquaux, G., Gramfort, A., et~al.
\newblock Scikit-learn: {{Machine Learning}} in {{Python}}.
\newblock \emph{Journal of Machine Learning Research}, 12\penalty0
  (85):\penalty0 2825--2830, 2011.

\bibitem[Ruppert(2004)]{ruppert2004elements}
Ruppert, D.
\newblock The {{Elements}} of {{Statistical Learning}}: {{Data Mining}},
  {{Inference}}, and {{Prediction}}.
\newblock \emph{Journal of the American Statistical Association}, 99\penalty0
  (466):\penalty0 567--567, June 2004.

\bibitem[Saeys et~al.(2007)Saeys, Inza, and Larranaga]{saeys2007review}
Saeys, Y., Inza, I., and Larranaga, P.
\newblock A review of feature selection techniques in bioinformatics.
\newblock \emph{Bioinformatics}, 23\penalty0 (19):\penalty0 2507--2517, October
  2007.

\bibitem[Saeys et~al.(2008)Saeys, Abeel, and {Van de Peer}]{saeys2008robust}
Saeys, Y., Abeel, T., and {Van de Peer}, Y.
\newblock Robust {{Feature Selection Using Ensemble Feature Selection
  Techniques}}.
\newblock In Daelemans, W., Goethals, B., and Morik, K. (eds.), \emph{Machine
  {{Learning}} and {{Knowledge Discovery}} in {{Databases}}}, volume 5212, pp.\
   313--325. {Springer Berlin Heidelberg}, {Berlin, Heidelberg}, 2008.
\newblock ISBN 978-3-540-87480-5 978-3-540-87481-2.

\bibitem[{Solorio-Fern{\'a}ndez} et~al.(2022){Solorio-Fern{\'a}ndez},
  {Carrasco-Ochoa}, and {Mart{\'i}nez-Trinidad}]{solorio-fernandez2022survey}
{Solorio-Fern{\'a}ndez}, S., {Carrasco-Ochoa}, J.~A., and
  {Mart{\'i}nez-Trinidad}, J.~F.
\newblock A survey on feature selection methods for mixed data.
\newblock \emph{Artificial Intelligence Review}, 55\penalty0 (4):\penalty0
  2821--2846, April 2022.

\bibitem[Song et~al.(2012)Song, Smola, Gretton, Bedo, and
  Borgwardt]{song2012feature}
Song, L., Smola, A., Gretton, A., Bedo, J., and Borgwardt, K.
\newblock Feature {{Selection}} via {{Dependence Maximization}}.
\newblock \emph{Journal of Machine Learning Research}, 13\penalty0 (5), 2012.

\bibitem[Song \& Li(2021)Song and Li]{song2021variable}
Song, Z. and Li, J.
\newblock Variable selection with false discovery rate control in deep neural
  networks.
\newblock \emph{Nature Machine Intelligence}, 3\penalty0 (5):\penalty0
  426--433, March 2021.

\bibitem[Tibshirani(1996)]{tibshirani1996regression}
Tibshirani, R.
\newblock Regression {{Shrinkage}} and {{Selection Via}} the {{Lasso}}.
\newblock \emph{Journal of the Royal Statistical Society: Series B
  (Methodological)}, 58\penalty0 (1):\penalty0 267--288, January 1996.

\bibitem[Wasserman(2010)]{wasserman2010all}
Wasserman, L.
\newblock \emph{All of Statistics: A Concise Course in Statistical Inference}.
\newblock Springer Texts in Statistics. {Springer}, {New York Berlin
  Heidelberg}, corr. 2. print., [repr.] edition, 2010.
\newblock ISBN 978-1-4419-2322-6.

\bibitem[Yamada et~al.(2014)Yamada, Jitkrittum, Sigal, Xing, and
  Sugiyama]{yamada2014highdimensional}
Yamada, M., Jitkrittum, W., Sigal, L., Xing, E.~P., and Sugiyama, M.
\newblock High-{{Dimensional Feature Selection}} by {{Feature-Wise Kernelized
  Lasso}}.
\newblock \emph{Neural Computation}, 26\penalty0 (1):\penalty0 185--207,
  January 2014.

\bibitem[Yang \& Moody(1999)Yang and Moody]{yang1999data}
Yang, H. and Moody, J.
\newblock Data {{Visualization}} and {{Feature Selection}}: {{New Algorithms}}
  for {{Nongaussian Data}}.
\newblock In \emph{Advances in {{Neural Information Processing Systems}}},
  volume~12. {MIT Press}, 1999.

\bibitem[Yang et~al.(2011)Yang, Shen, Ma, Huang, and Zhou]{yang2011l2}
Yang, Y., Shen, H.~T., Ma, Z., Huang, Z., and Zhou, X.
\newblock L2, 1-norm regularized discriminative feature selection for
  unsupervised.
\newblock In \emph{Twenty-Second International Joint Conference on Artificial
  Intelligence}, 2011.

\bibitem[Yang \& Pedersen(1997)Yang and Pedersen]{yang1997comparative}
Yang, Y. and Pedersen, J.~O.
\newblock A {{Comparative Study}} on {{Feature Selection}} in {{Text
  Categorization}}.
\newblock In \emph{Proceedings of the {{Fourteenth International Conference}}
  on {{Machine Learning}}}, {{ICML}} '97, pp.\  412--420, {San Francisco, CA,
  USA}, July 1997. {Morgan Kaufmann Publishers Inc.}
\newblock ISBN 978-1-55860-486-5.

\bibitem[Zhao et~al.(2013)Zhao, Wang, Liu, and Ye]{zhao2013similarity}
Zhao, Z., Wang, L., Liu, H., and Ye, J.
\newblock On {{Similarity Preserving Feature Selection}}.
\newblock \emph{IEEE Transactions on Knowledge and Data Engineering},
  25\penalty0 (3):\penalty0 619--632, March 2013.

\end{thebibliography}
	
	\bibliographystyle{nips-no-url}

	\newpage

\end{document}